\documentclass[]{acmart}

\usepackage{booktabs}
\usepackage{caption}
\usepackage{subcaption}
\usepackage{multirow}
\usepackage{enumitem}
\usepackage{fmtcount}
\usepackage{microtype}
\usepackage{balance}

\usepackage[ruled]{algorithm2e} 

\SetAlFnt{\small}
\SetAlCapFnt{\small}
\SetAlCapNameFnt{\small}
\SetAlCapHSkip{0pt}

\usepackage{ifthen}

\newcommand{\warning}[1]{{\it\color{red} #1}}
\newcommand{\toremove}[1]{{\it\color{red} (To remove) #1}}
\newcommand{\note}[1]{{\it\color{blue} #1}}
\newcommand{\nothing}[1]{}
\newcommand{\revision}[1]{{\color{black} #1}}

\definecolor{FanColor}{rgb}{0.8,0,0.8}

{

\renewcommand{\warning}[1]{}
\renewcommand{\toremove}[1]{}
\renewcommand{\note}[1]{}
\renewcommand{\nothing}[1]{}
}{}

\newcommand{\image}{I}
\newcommand{\inputcontent}{\image_c}
\newcommand{\inputstyle}{\image_s}

\newcommand{\augmentedimage}{\image^{+}}
\newcommand{\negativeexample}{\image^{-}}

\newcommand{\loss}{\mathcal{L}}
\newcommand{\latentcode}{\mathbf{z}}
\newcommand{\stylelatent}{\hat{\latentcode}}
\newcommand{\outputlatent}{\tilde{\latentcode}}

\newcommand{\negativelatent}{\mathbf{z_{i_j}^-}}
\newcommand{\positivepair}{\mathbf{s_i^+}}
\newcommand{\negativepair}{\mathbf{s_{i_j}^-}}
\newcommand{\positivetemperature}{\tau^+}
\newcommand{\negativetemperature}{\tau^-}

\newcommand{\generator}{G}

\AtBeginDocument{%
  \providecommand\BibTeX{{%
    \normalfont B\kern-0.5em{\scshape i\kern-0.25em b}\kern-0.8em\TeX}}}

\begin{document}

\title{A Unified Arbitrary Style Transfer Framework via Adaptive Contrastive Learning}


\author{Yuxin Zhang}
 \affiliation{
 \institution{NLPR, Institute of Automation, CAS}
 \country{China}
 }
 \affiliation{
 \institution{School of Artificial Intelligence, UCAS}
 \country{China}
 }
  \email{zhangyuxin2020@ia.ac.cn}

 \author{Fan Tang}
 \affiliation{
 \institution{Institute Of Computing Technology, CAS}
  \country{China}
 }
 \email{tangfan@ict.ac.cn}

 \author{Weiming Dong}
 \affiliation{
 \institution{NLPR, Institute of Automation, CAS}
 \country{China}
 }
 \affiliation{
 \institution{Peng Cheng Laboratory}
 \country{China}
 }
  \email{weiming.dong@ia.ac.cn}

 \author{Haibin Huang}
 \affiliation{
 \institution{Kuaishou Technology}
 \country{China}
 }
\email{huanghaibin03@kuaishou.com}

 \author{Chongyang Ma}
 \affiliation{
 \institution{Kuaishou Technology}
 \country{China}
 }
 \email{chongyangma@kuaishou.com}
  \author{Tong-Yee Lee}
 \affiliation{
 \institution{National Cheng-Kung University}
 \country{Taiwan}
 }
  \email{tonylee@ncku.edu.tw}

 \author{Changsheng Xu}
 \affiliation{
 \institution{NLPR, Institute of Automation, CAS}
 \country{China}
 }
 \affiliation{
 \institution{School of Artificial Intelligence, UCAS}
 \country{China}
 }
  \email{csxu@nlpr.ia.ac.cn}
\renewcommand{\shortauthors}{Zhang et al.}

\begin{abstract}
We present Unified Contrastive Arbitrary Style Transfer (UCAST), a novel style representation learning and transfer framework, which can fit in most existing arbitrary image style transfer models, e.g., CNN-based, ViT-based, and flow-based methods.
As the key component in image style transfer tasks, a suitable style representation is essential to achieve satisfactory results.
Existing approaches based on deep neural network typically use second-order statistics to generate the output.
However, these hand-crafted features computed from a single image cannot leverage style information sufficiently, which leads to artifacts such as local distortions and style inconsistency. 
To address these issues, we propose to learn style representation directly from a large amount of images based on contrastive learning, by taking the relationships between specific styles and the holistic style distribution into account.
Specifically, we present an adaptive contrastive learning scheme for style transfer by introducing an input-dependent temperature. 
Our framework consists of three key components, i.e., a parallel contrastive learning scheme for style representation and style transfer, a domain enhancement module for effective learning of style distribution, and a generative network for style transfer.
We carry out qualitative and quantitative evaluations to show that our approach produces superior results than those obtained via state-of-the-art methods.
\end{abstract}

% <concept_id>10010147.10010371.10010382.10010383</concept_id>
% <concept_desc>Computing methodologies~Image processing</concept_desc>




\begin{teaserfigure}
\centering
\includegraphics[width=\linewidth]{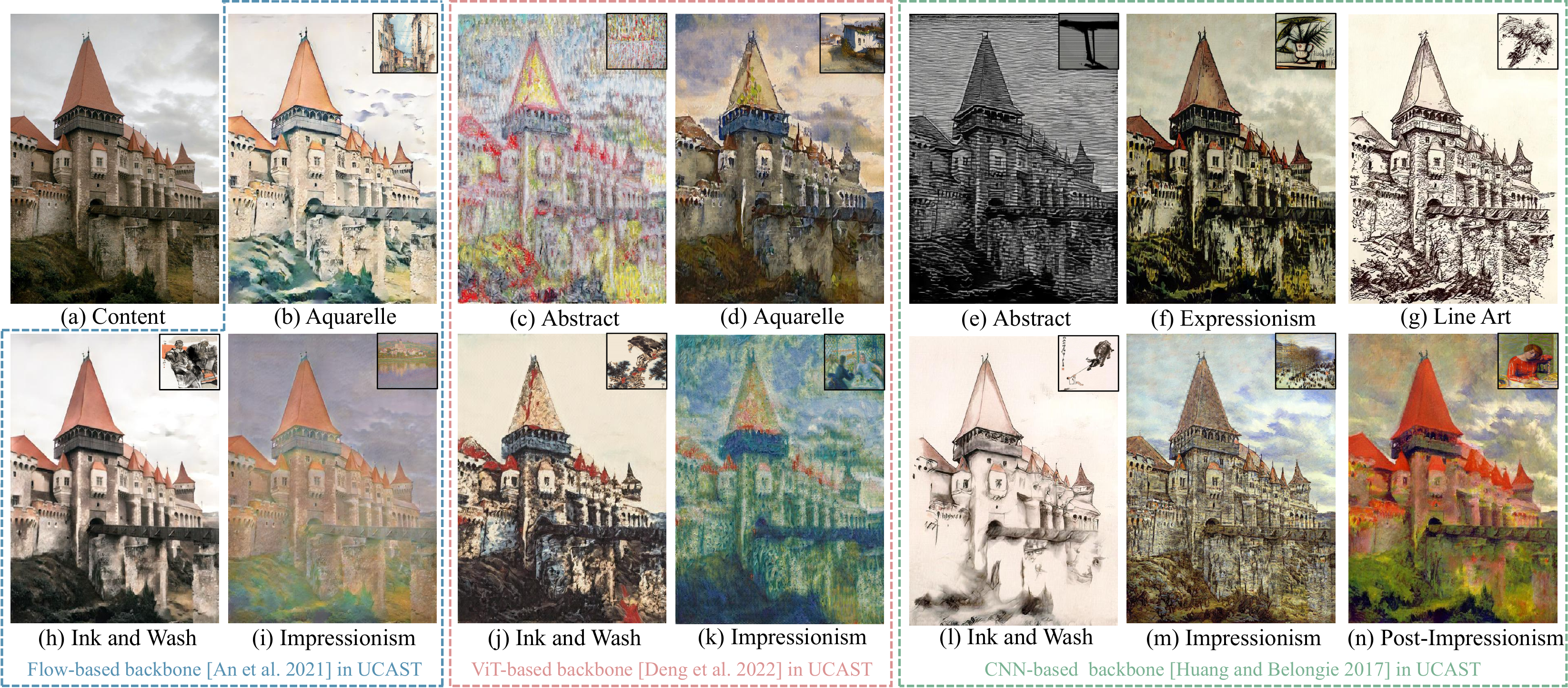}
\caption{Style transfer results of three different generative backbones trained under our framework, which can robustly and effectively handle various painting styles.
The input content image is shown in (a).
The style reference is shown as the inset for each result.
Our method can faithfully capture the style of each painting and generate a result with a unique artistic visual appearance.
}
\label{fig:teaser}
\end{teaserfigure}

\maketitle

%%%% 1-Introduction.tex starts here %%%%

\section{Introduction}
\label{sec:intro}

%%%% Figs/fig_IECAST_compare.tex starts here %%%%
\newcommand{\castiestfigurewidth}{0.158}

\begin{figure}%[t]
\centering
\includegraphics[width=0.99\linewidth]{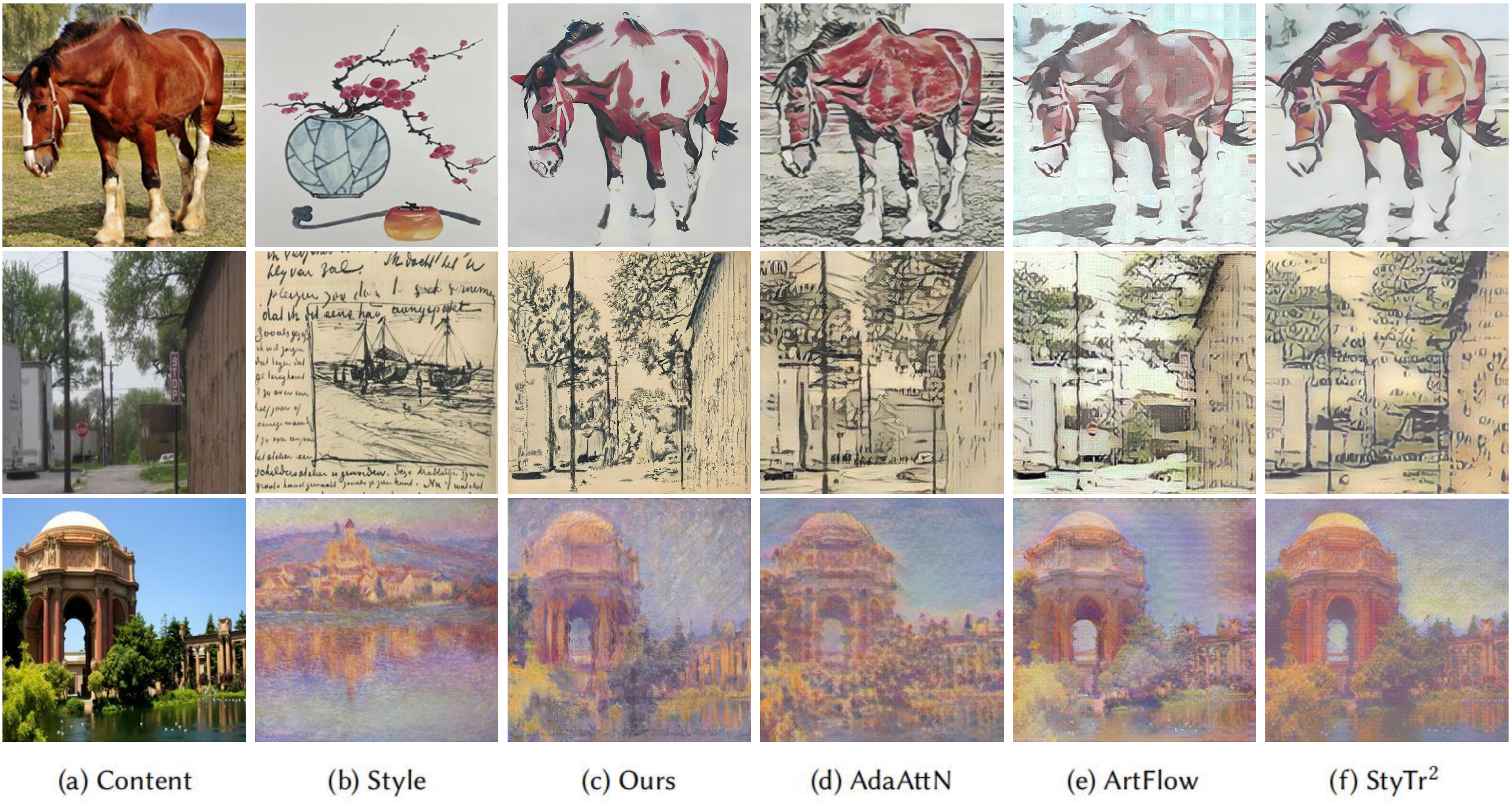}
\caption{Comparison with the latest style transfer methods: CNN-based method AdaAttN~\cite{liu2021adaattn}, neural flow-based method ArtFlow~\cite{An:2021:Artflow} and ViT-based method StyTr$^2$~\cite{deng2021stytr2}, all of which rely on second-order statistics.
Our method can faithfully transfer styles while ensuring structural consistency with the content images. 
}
\label{fig:motivation}
\end{figure}
%%%% Figs/fig_IECAST_compare.tex ends here %%%%

If a picture is worth a thousand words, then an artwork may tell the whole story.
The art style depicts the visual appearance of an artwork and characterizes how the artist expresses a theme and shows his/her creativity.
The features that identify an artwork, such as the artist's use of stroke, color and composition, determine the style.
Artistic style transfer, as an efficient way to create a new painting by combining the content of a natural images and the style of an existing painting image, is a major  research topic in computer graphics and computer vision~\cite{Jing:2020:NSTReview}.

The main challenges of arbitrary style transfer are extracting styles from artistic images and mapping a specific realistic image into an artistic one in a controllable way.
The core problem for style extraction is to find an effective representation of styles since it is in general hard to give explicit definitions across different styles. 
To build a reasonable style feature space, it is necessary to explore the relationship and distribution of styles in order to capture both individual and holistic characteristics.
For the mapping process, several generative mechanisms are adopted to address different issues, such as auto-encoder ~\cite{Huang:2017:AdaIn, liu2021adaattn}, neural flow model~\cite{An:2021:Artflow} and visual transformer~\cite{deng2021stytr2}.
In contrast to the goal of those methods, we propose to improve arbitrary style transfer via a unified framework that offers the guidance of proper artistic style representation and works for various generative backbones.

Since Gatys et al.~\shortcite{Gatys:2016:IST} proposed to use Gram matrix as artistic style representation, high-quality visual results are generated by advanced neural style transfer networks.
Despite the remarkable progress made in the field of arbitrary image style transfer, the second-order feature statistics (Gram matrix or mean/variance) style representation has restricted the further development and application.
As shown in Fig.~\ref{fig:teaser}, the appearances of different artwork styles vary considerably in terms of not only the colors and local textures but also the layouts and compositions.
Figs.~\ref{fig:motivation_adattn}-\ref{fig:motivation_stytr2} show the results of three recently proposed state-of-the-art style transfer approaches.
We obverse that aligning the distributions of neural activation between images using second-order statistics results in difficulty to capture the color distribution or the special layouts, or imitate specific detailed brush effects of different styles.

In this paper, we revisit the core problem for neural style transfer, that is, the proper artistic style representation.
The widely used second-order statistics as a global style descriptor can distinguish styles to some extent, but they are not the optimal way to represent styles.
By second-order statistics, arbitrary stylization formulates styles through artificially designed image features and loss functions in a heuristic manner.
In other words, the network learns to fit the second-order statistics of the style image and generated image, instead of the style itself.
Our key insight is that a person without artistic knowledge has difficulty defining the style if only one artistic image is given, but identifying the difference between different styles is relatively easy.
Therefore, exploring the relationship and distribution of styles directly from artistic images instead of using pre-defined style representations is worthwhile. 

We propose to improve arbitrary style transfer with a novel style representation based on contrastive learning.
Specifically, we present a \textbf{U}nified \textbf{C}ontrastive \textbf{A}rbitrary \textbf{S}tyle \textbf{T}ransfer (\textbf{UCAST}) framework for image style representation learning and style transfer, which consists of a generative backbone, a parallel contrastive learning scheme, and a domain enhancement (\textbf{DE}) module.
We introduce contrastive learning to consider the positive and negative relationships between different styles, and we use DE to learn the overall domain distribution of artistic images.
UCAST can be plug-and-play for most arbitrary image style transfer methods to improve their performance.

Since different images may share similar styles, it is necessary to consider similar styles in the style modeling process and the style contrastive learning should be tolerant to highly similar samples.
Besides, compared with Per-Style-Per-Model methods and Multiple-Style-Per-Model tasks, arbitrary image style transfer has the difficulty that when dealing with specific content-style pairs, the content image and style image may not always be compatible with each other.
For instance, when using a realistic image with a large smooth area as the content and an artistic image with rich texture as the style, we may find undesirable artifacts in the stylization output.
Based on these observations, we propose an adaptive contrastive loss that is implemented with a novel dual input-dependent temperature scheme.
Our adaptive contrastive loss takes into account the similarities between the target style image and other artistic images, as well as the similarities between the target style image and the input content image to address the above problems.

Our contribution can be summarized as follows:
\begin{itemize}[leftmargin=*,topsep=1pt]
\setlength\itemsep{0pt}
\item We propose a novel framework called Unified Contrastive Arbitrary Style Transfer (UCAST), which can easily integrate various types of style transfer backbones and lead to improved visual quality in the stylization results.

\item We propose a novel style representation learning method via contrastive learning without employing the commonly used second-order statistics of image features.
We introduce contrastive learning and domain enhancement by considering the relationships between styles as well as the global distribution of styles, which solves the problem that existing style transfer models cannot effectively leverage a large amount of available artistic images. 

\item We propose adaptive contrastive learning for arbitrary style transfer tasks, which allows the model to be tolerant to similar styles and improve the robustness of various content-style inputs.

\end{itemize}

%%%% Figs/fig_framework.tex starts here %%%%

\begin{figure*}%[t]
\centering
\includegraphics[width=0.99\linewidth]{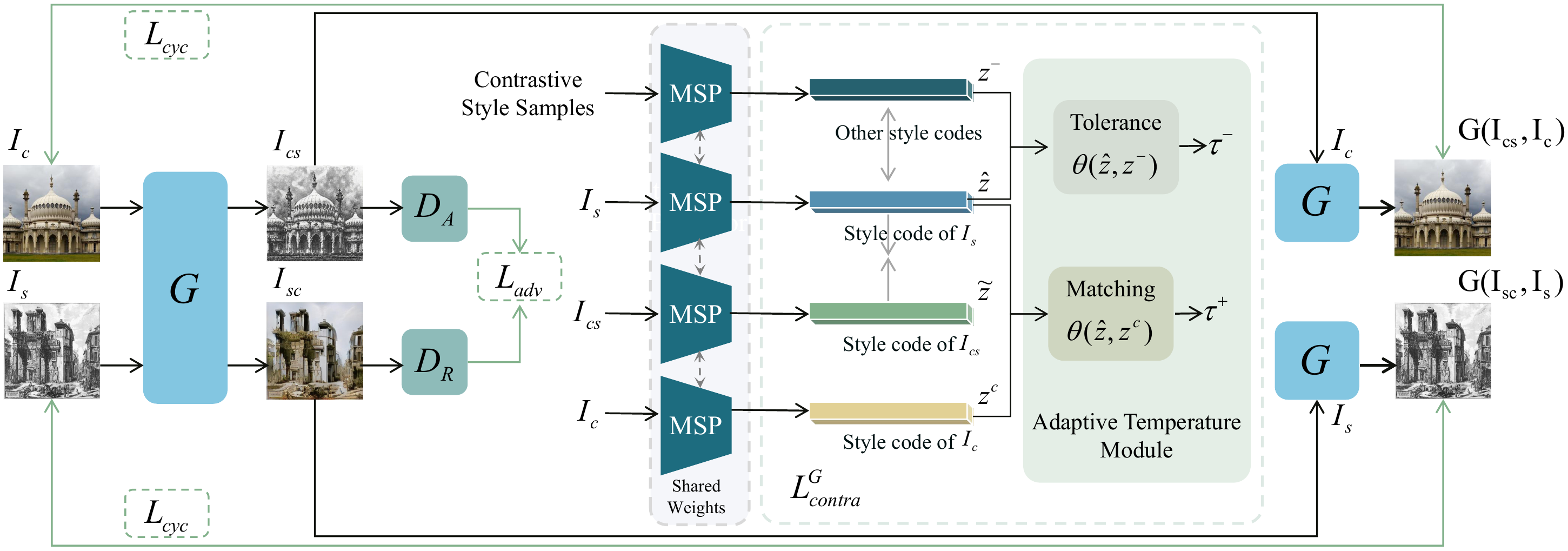}
%\vspace{-6pt}
\caption{UCAST consists of an generator $\generator$, a parallel contrastive learning scheme relying on a multi-layer style projector (MSP) module, and a domain enhancement module.
The generator is given the content image $\inputcontent$ and the style image $\inputstyle$ and generates images $\image_{cs}$ and $\image_{sc}$.
Then, $\image_{cs}$ and $\inputstyle$ are fed into the MSP module to generate the corresponding style code $\outputlatent$ and $\stylelatent$, which will be used as positive samples in the style contrastive learning process. 
The style codes $\latentcode^{-}$ of other artistic images in the style bank will be used as negative samples.
$\inputcontent$ is fed into the MSP module to generate the corresponding style code $\latentcode^c$.
We design an adaptive temperature module that computes the temperature $\positivetemperature$ of the positive sample and the temperature $\negativetemperature$ of the negative samples on the style codes.
The contrastive style loss $\loss_{contra}^{G}$ is computed on the temperatures and the style codes.
DE module is based on the adversarial loss $\loss_{adv}$ and the cycle consistency loss $\loss_{cyc}$.
}
%\vskip -2.5mm
\label{fig:framework}
\end{figure*}

%%%% Figs/fig_framework.tex ends here %%%%

%%%% 1-Introduction.tex ends here %%%%
%%%% 2-RelatedWork.tex starts here %%%%

\section{Related Work}

\paragraph{Image style transfer.}
Traditional style transfer methods such as stroke-based rendering~\cite{Fivser:2016:Stylit} and image filtering~\cite{Wang:2004:EEP} typically use low-level hand-crafted features.
Gatys et al.~\shortcite{Gatys:2016:IST} and the follow-up variants \cite{Gatys:2017:CPF,Kolkin:2019:STR} demonstrate that the statistical distribution of features extracted from pre-trained deep convolutional neural networks can capture style patterns effectively.
Although the results are remarkable, these methods formulate the task as a complex optimization problem, which leads to high computational cost.
Some recent approaches rely on a learnable neural network to match the statistical information in feature space for efficiency.
Per-Style-Per-Model methods~\cite{johnson2016perceptual,gao2020fast, puy2019flexible,Kwon:2022:clipstyler} train a specific network for each individual style.
Multiple-Style-Per-Model methods~\cite{chen2017stylebank,zhang2018multi,dumoulin2016learned, ulyanov2016texture} represent multiple styles using one single model.

Arbitrary style transfer methods~\cite{Liao:2017:VAT,Li:2017:UST,deng2020arbitrary,svoboda2020two,Wu:2021:SF,deng2021stytr2,zhang2020cast} build more flexible feed-forward architectures to handle an arbitrary style using a unified model.
AdaIN~\cite{Huang:2017:AdaIn} and DIN~\cite{jing2020dynamic} directly align the overall statistics of content features with the statistics of style features and adopt conditional instance normalization.
However, dynamic generation of affine parameters in the instance normalization layer may cause distortion artifacts.
Instead, several methods follow the encoder-decoder manner, where feature transformation and/or fusion is introduced into an auto-encoder-based framework. 
For instance, Li et al.~\shortcite{Li:2019:LLT} achieve universal style transfer by developing a cross-domain feature linear transformation matrix (LST) and decoding from the transformed features.
Park et al.~\shortcite{Park:2019:AST} provide a flexible mapping of the semantically nearest style features onto the content features by SANet.
Deng et al.~\shortcite{Deng:2021:MCC} propose MCCNet for efficient video style transfer by fusing input content features and style features via multi-channel correlation. 
Liu et al.~\shortcite{liu2021adaattn} present an adaptive attention normalization module (AdaAttN) to consider both shallow and deep features for attention score calculation.
GAN-based methods~\cite{Zhu:2017:CycleGAN,svoboda2020two, kotovenko2019content_transformation, kotovenko2019content, sanakoyeu2018style} have been successfully used in collection style transfer, which considers style images in a collection as a domain~\cite{chen2021dualast, xu2021drb,Lin:2021:DAM}.
An et al.~\shortcite{An:2021:Artflow} propose reversible neural flows and an unbiased feature transfer module (ArtFlow) to prevent content leak during universal style transfer.
Inspired by the breakthrough of visual transformer (ViT), many researchers have developed ViT for style transfer tasks.
Wu el al.~\shortcite{Wu:2021:SF} propose a feed-forward style transfer method (StyleFormer) which includes a transformer-driven style composition module.
Deng et al.~\shortcite{deng2021stytr2} propose a ViT-based style transfer method (StyTr$^2$) which takes long-range dependencies of input images into account to avoid the biased content representation.
Zhang et al.~\shortcite{Zhang:2022:EFD} performed exact matching of feature distributions and applied this method to arbitrary style transfer.

\paragraph{Contrastive learning.}
Contrastive learning has been used in many applications, such as image dehazing~\cite{Wu:2021:CLC}, context prediction~\cite{Cruz:2019:VPL}, geometric prediction~\cite{Liu:2019:EUD} and image translation.
Contrastive learning is introduced in image translation to preserve the content of the input~\cite{Han:2021:DCL} and reduce mode collapse~\cite{Liu:2021:DivCo,Jeong:2021:Contrad,Kang:2020:ContraGAN}.
CUT~\cite{park2020CUT} proposes patch-wise contrastive learning by cropping input and output images into patches and maximizing the mutual information between patches.
Following CUT, TUNIT~\cite{baek2021tunit} adopts contrastive learning on images with similar semantic structures.
However, the semantic similarity assumption does not hold for arbitrary style transfer tasks, which leads the learned style representations to a significant performance drop.
IEST~\cite{chen2021artistic} applies contrastive learning to image style transfer based on feature statistics (mean and standard deviation) as style priors.
The contrastive loss is calculated only within the generated results.
Contrastive learning in IEST is an auxiliary method to associate stylized images sharing the same style, and the ability comes from the feature statistics from pre-trained VGG.
Differently, we introduce contrastive learning for style representation by proposing a novel framework that uses visual features comprehensively to represent style for the task of arbitrary image style transfer.

Temperature is a critical parameter for the success of a contrastive learning based method.
Wang and Liu~\shortcite{wang2021understanding} showed that the contrastive loss has a hardness-aware property, which makes contrastive learning naturally focusing on difficult negative samples.
Such hardness awareness helps to learn separable and uniformly distributed features but also leads to low tolerance of semantically similar samples.
The extent of penalties on hard negative samples is determined by the temperature $\tau$.
In particular, as the temperature decreases, the relative penalty tends to concentrates more on the high similarity region, whereas as the temperature increases, the relative penalty distribution becomes more uniform, which means that all negative samples are penalized equally.
They built relations between uniformity, tolerance, and temperature.
Zhang et al.~\shortcite{zhang2022does} introduced vector decomposition for analyzing the collapse issue based on gradient analysis of the $l_2$-normalized representation vector and proposed a unified perspective on how negative samples and simple Siamese method alleviate collapse.
Caron et al.~\shortcite{caron2021emerging} investigated dual temperature from the perspective of knowledge distillation and proposed a simple self-supervised method, in which the teacher adopts a lower temperature than the student to help in knowledge distillation.
Zhang et al.~\shortcite{zhang2021temperature} learn temperature as an input-dependent variable.
They consider temperature as a measure of embedding confidence and propose temperature as uncertainty.
Zhang et al.~\shortcite{zhang2022dual} adopt dual temperature in a contrastive InfoNCE for realizing independent control of two hardness-aware sensitiveness.
Previous temperature analysis works mainly focus on the penalty's unevenness of negative samples within an anchor or the sum of penalties of different anchors within a training batch.
Contrarily, we consider the proportion of penalties between the positive sample and negative samples at the same time.
%%%% 2-RelatedWork.tex ends here %%%%

%%%% 3-Method.tex starts here %%%%

\section{Method}

\subsection{Overview}
Our unified framework for arbitrary image style transfer as a separated network structure can be plug-and-play for most arbitrary image style transfer models.
As shown in Fig.~\ref{fig:framework}, our UCAST consists of three key components:
1) a parallel contrastive learning scheme that is applied to the style representation learning and the style transfer process; 
2) a domain enhancement scheme to further help learn the distribution of the artistic image domain;
and 3) a generator $\generator$ to generate the stylization output.
Both 1) and 2) are used for learning style features to measure the difference between artistic images and realistic images.
The parallel contrastive learning scheme focuses on forcing the specific reference artistic image and the generated result to have the same style, while the domain enhancement scheme pays attention to the holistic difference between the artistic domain and the realistic domain.

The main structure of our parallel contrastive learning scheme is a multi-layer style projector that is trained to project features of artistic images into style codes.
The contrastive losses are introduced to guide parallel optimization processes, including both the training of  multi-layer style projector and the generator.
When training the generator, we introduce adaptive contrastive loss which is implemented with dual input-dependent temperature.
By considering the similarities between the style codes of the reference style image and other artistic images, our adaptive contrastive loss is more tolerant to style consistent samples.
Meanwhile, the input-dependent temperature is also influenced by the similarities between the style codes of the target style image and the input content image, to increase the robustness of various content-style pairs and prevent artifacts.
The domain enhancement scheme is accomplished by two discriminators for the artistic domain and the realistic domain respectively.
The adversarial loss helps the discriminator model the distribution of the corresponding domain and the cycle consistency loss is adopted to maintain the content information.

\begin{figure*}
\centering
\includegraphics[width=0.99\linewidth]{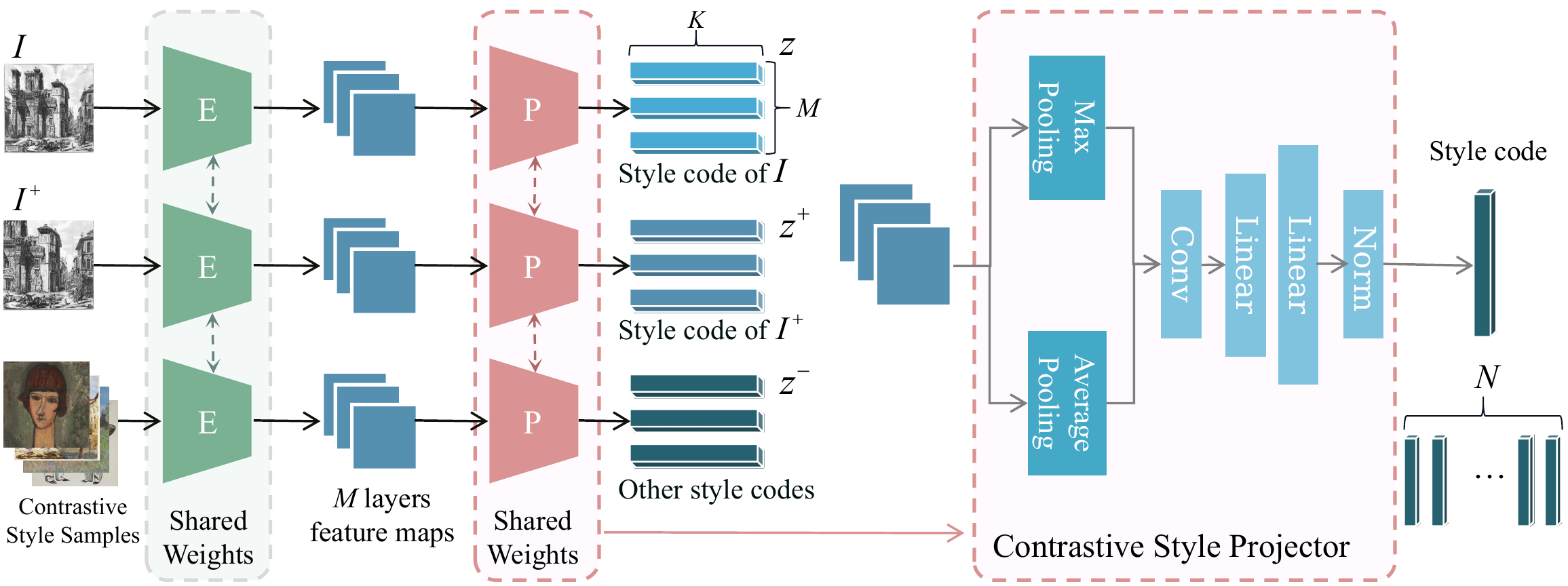}
\caption{Overview of our MSP module, which includes a VGG-19 based style feature extractor $E$ and a multi-layer projector $P$. 
$P$ maps the extracted features to style codes $\{ \latentcode \}$ which will be then saved in the style memory bank.
}
\label{fig:msp_overview}
\end{figure*}

\subsection{Parallel Contrastive Learning}

\subsubsection{Multi-layer Style Projector}

Our goal is to develop an unified arbitrary style transfer framework that can capture and transfer the local stroke characteristics and overall appearance of an artistic image to a natural image.
A key component is to find a suitable style representation which can be used to distinguish different styles and further guide the generation of style images.
To this end, we design an MSP module, which includes a style feature extractor and a multi-layer projector.
Instead of using features from a specific layer or a fusion of multiple layers, our MSP projects features of different layers into separate latent style spaces to encode local and global style cues.

Specifically, we adopt VGG-19~\cite{simonyan2014very} and fine-tune the VGG-19 model pre-trained on ImageNet with a collection of 18,000 artistic images in 50 categories.
We then select $M$ layers of feature maps in VGG-19 as input to our multi-layer projector (we use layers of ReLU1\_2, ReLU2\_2, ReLU3\_3, and ReLU4\_3 in all experiments).
We use max pooling and average pooling to capture the mean and peak value of features.
The multi-layer projector consists of pooling, convolution, and several multilayer perceptron layers, and it projects the style features into a set of $K$-dimensional latent style code, as shown in Fig.~\ref{fig:msp_overview}.

After training, MSP can encode an artistic image into a set of latent style code $\{ \latentcode_i | i \in [1, M], \latentcode_i \in \mathbb{R}^K\}$, which can be plugged into an existing style transfer network (i.e., replacing the mean and variance in AdaIN~\cite{Huang:2017:AdaIn}) as the guidance for stylization.
Next, we will describe how to jointly train MSP and style transfer networks with a contrastive learning strategy.

\subsubsection{Contrastive Style Representation Learning}

A branch of the parallel contrastive learning scheme is the style representation learning.
The MSP needs to be trained in order to obtain the reasonable style representation which is in the form of the style code $\{ \latentcode_1, \latentcode_2,..., \latentcode_M \}$.
However, we lack the ground-truth style code for supervised training.
Therefore, we adopt contrastive learning and design a new contrastive style  loss as an implicit measurement for the MSP training.

When training the MSP module, an image $\image$ and its augmented version $ \augmentedimage $ (random resizing, cropping, and rotations) are fed into a $M$-layer style feature extractor, which is the pre-trained VGG-19 network.
The extracted style features are then sent to the multi-layer projector, which is an $M$-layer neural network and maps the style features to a set of $K$-dimensional vectors $\{ \latentcode \}$.
The contrastive representation learns the visual styles of images by maximizing the mutual information between $\image$ and $\augmentedimage$ in contrast to other artistic images within the dataset considered as negative samples $\{ \negativeexample \}$.
Specifically, the images $\image$, $\augmentedimage$, and $N$ negative samples are respectively mapped into $M$ groups of $K$-dimensional vectors $\latentcode$, $\latentcode^{+} \in \mathbb{R}^K$ and $\{ \latentcode^{-} \in \mathbb{R}^K \}$. The vectors are normalized to prevent collapsing.
We maintain a large dictionary of 4096 negative examples using a memory bank architecture following MOCO~\cite{He:2019:MOCO}.
The negative examples are sampled from the memory bank.
Following \cite{van2018representation}, we define the contrastive loss function to train our MSP module as:
\begin{equation}
\begin{aligned}
\loss_{contra}^{MSP}=-\sum_{i=1}^M{\log \frac{\exp(\latentcode_{i} \cdot {\latentcode_{i}^{+}}/ \tau)}{\exp (\latentcode_{i} \cdot {\latentcode_{i}^{+}} / \tau)+\sum_{j=1}^N{\exp(\latentcode_{i} \cdot {\latentcode_{i_{j}}^{-}} / \tau)}}},
\end{aligned}
\label{eqn:loss_msp}
\end{equation}
where $\cdot$ denotes the dot product of two vectors.
It is worth noting that we calculate the contrastive loss between \emph{images}, as opposed to CUT~\cite{park2020CUT} which adopts contrastive learning by cropping images into patches and maximizing the mutual information between \emph{patches}.

%%%% Figs/fig_adaptive_temperature.tex starts here %%%%

\begin{figure*}
\centering
\begin{subfigure}[t]{0.32\linewidth}
\includegraphics[width=1\linewidth]{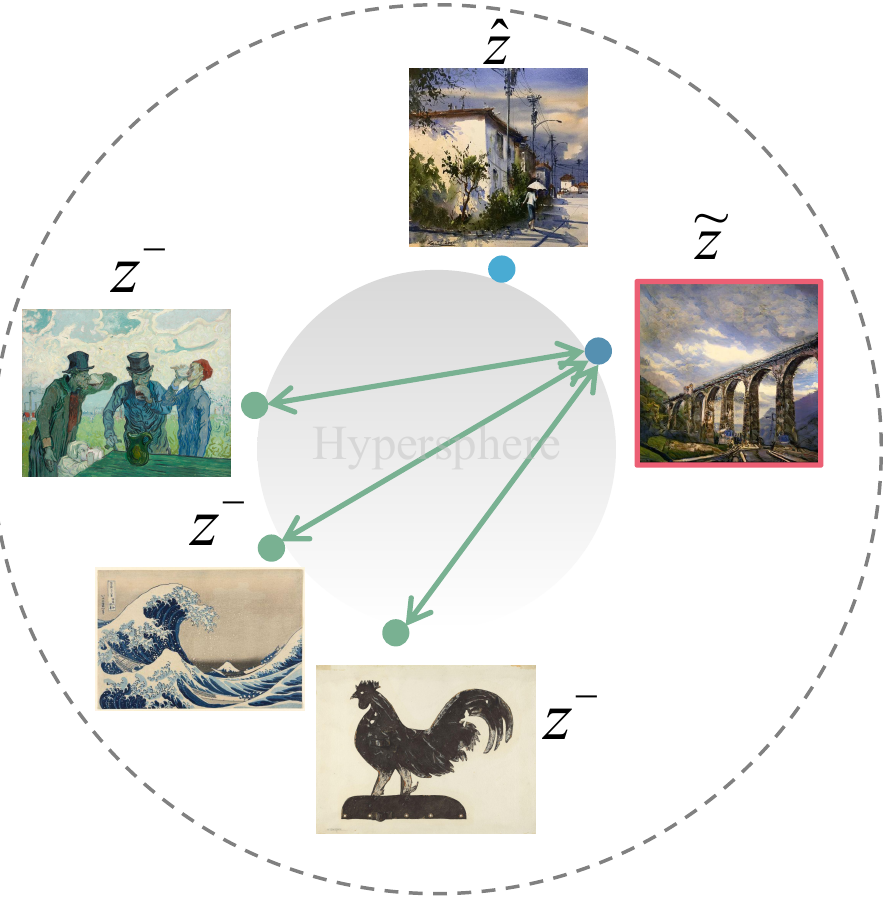}
\caption{Fixed temperature: Case I}
\end{subfigure}
\begin{subfigure}[t]{0.32\linewidth}
\includegraphics[width=1\linewidth]{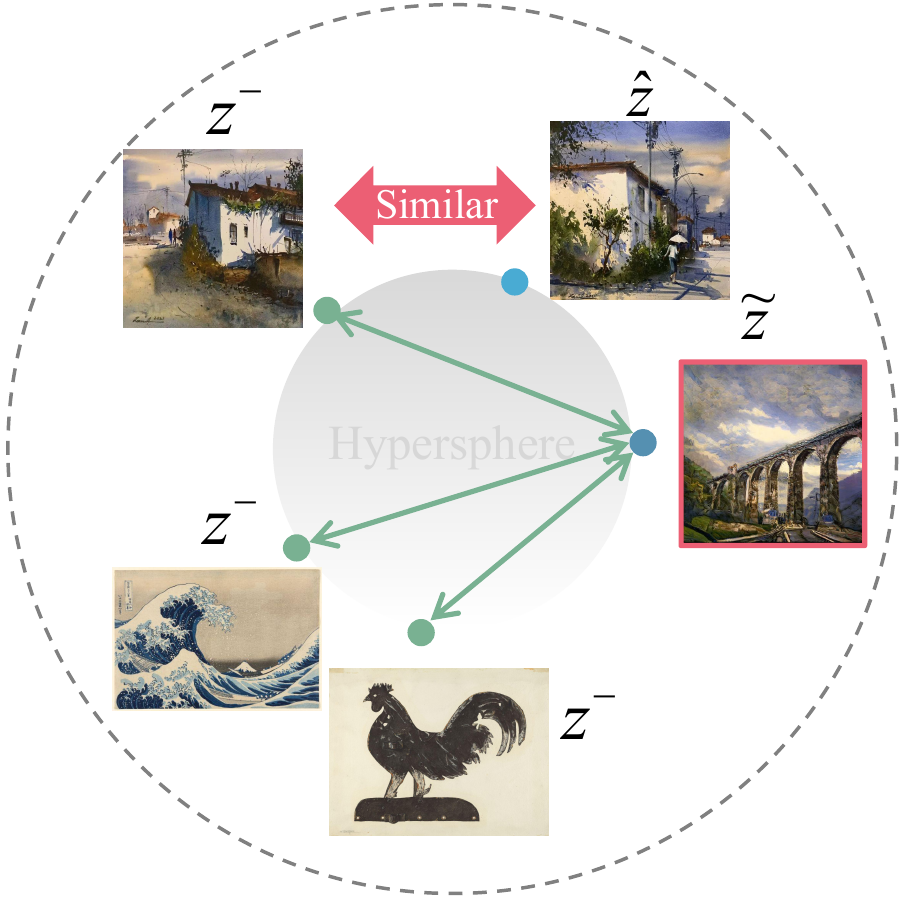}
\caption{Fixed temperature: Case II}
\end{subfigure}
\begin{subfigure}[t]{0.32\linewidth}
\includegraphics[width=1\linewidth]{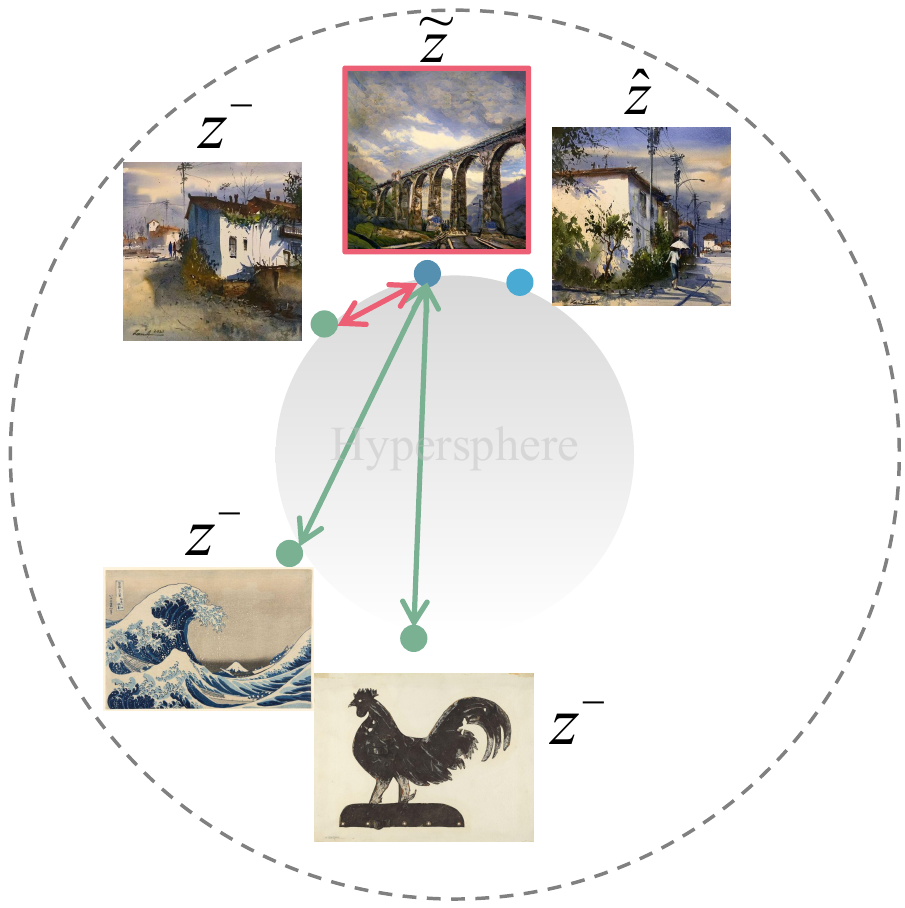}
\caption{Adaptive temperature: Case II}
\end{subfigure}
\caption{\revision{
Visualization of the embedding distribution of artistic images and generated results on a hypersphere. 
}}
\label{fig:adaptive_temperature}
\end{figure*}

%%%% Figs/fig_adaptive_temperature.tex ends here %%%%

\subsubsection{Contrastive Style Transfer}

The other branch of the parallel contrastive learning scheme is the style transfer process.
The above contrastive representation provides proper measurement for the generator $\generator$ to transfer styles between images.
We compute the loss using the contrastive representations of the output image $\image_{cs}$ and the reference style image $\inputstyle$, then $\image_{cs}$ will have a style similar to $\inputstyle$:
\begin{equation}
\begin{aligned}
\loss_{contra}^{G}=-\sum_{i=1}^M {\log \frac{\exp({\outputlatent}_i \cdot {\stylelatent}_i/ \tau)}{\exp ({\outputlatent}_i \cdot {\stylelatent}_i / \tau)+\sum_{j=1}^N{ \exp({\outputlatent}_i \cdot {\negativelatent} / \tau)}}},
\end{aligned}
\label{eqn:loss_G_NCE}
\end{equation}
where $\outputlatent$ and $\stylelatent$ denote the contrastive representation of $\image_{cs}$ and $\inputstyle$, respectively.
Notably, we take the specific generated and reference images as positive examples and utilize contrastive loss as guidance to transfer styles, which is an one-on-one process.
Differently, the contrastive loss in IEST~\cite{chen2021artistic} is calculated only within generated results and it takes a set of images as positive examples, which could reduce the style consistency with the given reference (see Fig.~\ref{fig:SOTA}).

\subsubsection{Adaptive Contrastive Learning}

Since different artworks could have similar styles, the model needs to be tolerant to these style similarities.
Contrastive learning seeks to minimize the distance between positive samples and maximize the distance between negative samples in the representation space.
By gradient analysis, \cite{wang2021understanding} demonstrates that the gradients with regard to negative samples are proportional to the similarity between the particular negative sample and the anchor, proving that the contrastive loss is a hardness-aware loss function.
The temperature $\tau$ controls the distribution of negative gradients. 
Smaller temperatures tend to focus more on the anchor point's nearest neighbors, whereas larger temperatures tend to penalize negative samples equally. 
When the temperature is fixed, the gradient's magnitude with respect to a positive sample is equal to the sum of gradients with respect to all negative samples.
Prior works of temperature analysis mainly focus on the penalty's unevenness of negative samples within an anchor~\cite{wang2021understanding}, or the sum of penalties of different anchors within a training batch~\cite{zhang2022dual}.
Differently, we pay attention to the proportion of penalties between the positive sample and negative samples.

In Fig.~\ref{fig:adaptive_temperature}, we show the embedding distribution with four real paintings and one generated image on a hypersphere. 
As shown in Fig.~\ref{fig:adaptive_temperature}a, when the style of the reference image and the other artistic images served as negative samples vary differently, the punishment of the fixed small temperature may work finely.
Different artistic images may share similar styles.
When dealing with the case that similar style images act as negative samples, as shown in Fig.~\ref{fig:adaptive_temperature}b, the ideal embedding of the generated image is being separated from all the negative samples but closer to the similar negative samples.
However, the contrastive loss with fixed small temperature tends to give strong punishment on similar samples due to the hardness-aware attribute, which means the generated image may be pushed away from the similar negative sample too much which is not a reasonable embedding in the hypersphere.
Our adaptive contrastive style transfer approach is aware of the negative samples which share a similar style with the reference image.
When high similarity negative samples appear, our approach will gain tolerance by increasing the temperature accordingly.
As shown in Fig.~\ref{fig:adaptive_temperature}c, with the help of our adaptive contrastive style transfer approach, the generator will be guided under a reasonable loss and the generated image can get a better embedding.

To further illustrate our adaptive temperature mechanism, we substitute the similarities of the positive sample and the negative samples in Eq.~\ref{eqn:loss_G_NCE} with $\positivepair =  \outputlatent_i \cdot \stylelatent,
\negativepair =  \outputlatent_i \cdot \negativelatent$:
\begin{equation}
\begin{aligned}
\loss_{contra}^{G}=-\sum_{i=1}^M {\log \frac{\exp(\positivepair/ \positivetemperature)}{\exp (\positivepair / \positivetemperature)+\sum_{j=1}^N{ \exp(\negativepair / \negativetemperature)}}},
\end{aligned}
\label{eqn:loss_G_NCE2}
\end{equation}
where $\positivetemperature$ and $\negativetemperature$ indicate the temperatures of positive samples and negative samples, respectively.
We analyze the gradients with respect to positive samples and different negative samples.
Specifically, the gradients with respect to the positive similarity $\positivepair$ and the negative similarity $\negativepair$ are formulated as:
\begin{equation}
\begin{aligned}
\frac{\partial \loss_{contra}^{G}}{\partial \positivepair} = - \sum_{i=1}^M \frac{1}{\positivetemperature} \cdot \frac{\sum_{j=1}^N{ \exp(\negativepair / \negativetemperature)}}{\exp (\positivepair / \positivetemperature)+\sum_{j=1}^N{ \exp(\negativepair / \negativetemperature)}},\\
\frac{\partial \loss_{contra}^{G}}{\partial \negativepair} = - \sum_{i=1}^M \frac{1}{\negativetemperature} \cdot \frac{\exp(\negativepair / \negativetemperature)}{\exp (\positivepair / \positivetemperature)+\sum_{j=1}^N{ \exp(\negativepair / \negativetemperature)}}.
\end{aligned}
\label{eqn:loss_G_partial}
\end{equation}
From Eq.~(\ref{eqn:loss_G_partial}), we can see that the magnitude of the gradient with respect to the positive sample is proportional to the sum of gradients with respect to all the negative samples.
By controlling $\negativetemperature$ and $\positivetemperature$, we can change the strength of penalties on the positive sample and negative samples.

We propose an input-dependent scheme to determine temperature by considering the similarities between the style code of the reference style $\stylelatent$ and the style codes of other artistic images $\negativelatent$.
Specifically, the more highly similar samples the memory bank contains, the larger the temperature is.
% Taking $\mu^-$ and $\sigma^-$ indicate the estimation of the mean and standard deviation of $\sum_{j=1}^N g(\negativepair)$.
Our input-dependent temperature is computed as:
\begin{equation}
\begin{aligned}
\negativetemperature &= t^-_{range} \cdot \frac{1}{1+\exp(-(\sum_{j=1}^N  g(\negativepair)-\mu^-)\cdot \sigma^-)} + t^-_{bound}, \\
g(\negativepair) &= \left\{ \begin{array}{rcl}
\negativepair & \mbox{for} & \negativepair > \mathbf{s^-}\\
0 & \mbox{for} & \negativepair \leq \mathbf{s^-}
\end{array} \right. ,
\end{aligned}
\label{eqn:loss_negative_temperature}
\end{equation}
where $\mu^-$ and $\sigma^-$ indicate the estimation of the mean and standard deviation of $\sum_{j=1}^N  g(\negativepair)$.
$t^-_{range}$ and $t^-_{bound}$ denote the range and lower bound of $\negativetemperature$.

Besides, arbitrary style transfer task often has the problem that the style images may not always be suitable for the content image and thus increase undesired artifacts. 
For example, when transferring a texture-rich style to a smooth content image, the model may produce artifacts and distortion (e.g., the 4\textsuperscript{th} row of Fig.~\ref{fig:SOTA}).
Therefore, it is necessary to adaptively handle various content-style pairs to increase the robustness.
To overcome the aforementioned problem, we propose a suitability-aware scheme to determine temperature based on the similarity between the style code of the reference image $\stylelatent$ and the style code of the content image $\latentcode_i^c$.
When the reference style and the content image are dissimilar, the penalty tends to be assigned more to negative samples to prevent artifacts of being overly stylized:
\begin{equation}
\begin{aligned}
\positivetemperature &= \negativetemperature \cdot f(\stylelatent,\latentcode_i^c), \\
f(\stylelatent,\latentcode_i^c) &= t^+_{range} \cdot \frac{1}{1+\exp((\stylelatent \cdot \latentcode_i^c -\mu^+) \cdot \sigma^+)} + t^+_{bound},
\end{aligned}
\label{eqn:loss_positive_temperature}
\end{equation}
where $\mu^+$ and $\sigma^+$ indicate the estimation of the mean and standard deviation of $\stylelatent \cdot \latentcode_i^c)$.
$t^+_{range}$ and $t^+_{bound}$ denote the range and lower bound of $\positivetemperature$.

\subsection{Domain Enhancement}
We introduce DE with adversarial loss to enable the network to learn the style distribution.
Recent style transfer models employ GAN~\cite{goodfellow2014generative} to align the distribution of generated images with specific artistic images~\cite{chen2021dualast,Lin:2021:DAM}.
The adversarial loss can enhance the holistic style of the stylization results, while it strongly relies on the distribution of datasets.
Even with the specific artistic style loss, the generation process is often not robust enough to be artifact-free.

Differently from these previous methods, we divide the images in the training set into realistic domain and artistic domain, and we use two discriminators $D_R$ and $D_A$ to enhance them respectively (see Fig.~\ref{fig:framework}).
During the training process, we first randomly select an image from the realistic domain as the content image $\inputcontent$ and another image from the artistic domain as the style image $\inputstyle$.
$\inputcontent$ and $\inputstyle$ are used as the real samples of $D_R$ and $D_A$, respectively.
The generated image $\image_{cs} = \generator(\inputcontent, \inputstyle)$ is used as the fake sample of $D_A$.
We exchange the content and style images to generate an image $\image_{sc} = \generator(\inputstyle, \inputcontent)$ as the fake sample of $D_R$. The adversarial loss is:
\begin{equation}
\begin{aligned}
\loss_{adv} =& \mathbb{E}[\log D_R(I_c)]+\mathbb{E}[\log (1-D_R(I_{cs}))]\\
& + \mathbb{E}[\log D_A(I_s)]+\mathbb{E}[\log (1-D_A(I_{sc}))].
\end{aligned}
\end{equation}

To maintain the content information of the content image in the process of style transfer between the two domains, we also add a cycle consistency loss:
\begin{equation}
\begin{aligned}
    \loss_{cyc} = \mathbb{E} [\Vert I_c -\generator(I_{cs},I_c)\Vert_1] + \mathbb{E} [\Vert I_s -\generator(I_{sc},I_s)\Vert_1].
\end{aligned}
\end{equation}

\subsection{Video Style Transfer}

To apply our method for video style transfer, we adopt the patch-wise contrastive content loss in~\cite{park2020CUT} to keep the content consistency.
The feature maps of the content image and the stylized result are cut into feature patches.
The patches at the same specific location of the content image and the stylized result are leveraged as positive samples, while the other patches within the input as negatives:
\begin{equation}
\begin{aligned}
\loss_{contra}^{c}=-{\log \frac{\exp(v \cdot {v^{+}}/ \tau)}{\exp (v \cdot {v^{+}} / \tau)+\sum_{n=1}^W{\exp(v \cdot {v_n^{-}} / \tau)}}},
\end{aligned}
\label{eqn:loss_cut}
\end{equation}
where $v,v^+ \in \mathbb{R}^K$, $v_n^- \in \mathbb{R}^{K \times W}$ denote the content feature of generated image patch, content image patch, and negative image patches, respectively.

\subsection{Network Training}
Our full objective function for training of the generator $G$ and discriminators $D_R$ and $D_A$ is formulated as:
\begin{equation}
\begin{aligned}
\loss(G, D_R, D_A) &= \lambda_{1} \loss_{adv}+ \lambda_{2}\loss_{cyc}+ \lambda_{3} \loss^G_{contra}+
\lambda_{4} \loss_{contra}^{c},\\
\end{aligned}
\label{eqn:total_loss}
\end{equation}
where $\lambda_{1}$, $\lambda_{2}$, $\lambda_{3}$, and $\lambda_{4}$ are weights to balance different loss terms. We set $\lambda_{1} = 1.0$, $\lambda_{2} = 2.0$, $\lambda_{3} = 0.2$, and $\lambda_{4} = 1.0$ in our experiments.

%%%% 3-Method.tex ends here %%%%

%%%% Tables/tab_quantitative.tex starts here %%%%

\begin{table*}%[h]
\centering
\caption{
Statistics of inference speed and quantitative comparison with state-of-the-art methods.
The results of user study \uppercase\expandafter{\romannumeral1} represent the average percentage of cases in which the result of the corresponding method is preferred compared with ours.
The results of user study \uppercase\expandafter{\romannumeral2} show the accuracy and recall of being selected as fake paintings by the participants.
The best results are in \textbf{bold} and the second best results are marked with \underline{underlines}.
}
%\resizebox{1\linewidth}{!}{%
\begin{tabular}{c||c|c|c|c|c|c|c}
\toprule
Method & Inference time & Content loss$\downarrow$ & LPIPS$\downarrow$ & Deception Rate$\uparrow$ & User Study  \uppercase\expandafter{\romannumeral1} & \multicolumn{2}{c}{User Study\uppercase\expandafter{\romannumeral2} } \\ 
\cline{7-8}
& (ms/image) &  &  &  &  & Precision$\downarrow$ & Recall$\downarrow$  \\ \hline \hline
StyleTr$^2$ & 87 & 0.123 & 0.311 & 54.7\% & 38.3\% & 59.0\% & 56.7\% \\ \hline
StyleFormer & \textbf{8} & 0.176 & 0.329 & 53.2\% & 39.6\% & 67.2\% & 63.4\% \\ \hline
IEST & 184 & 0.134 & 0.305 & 58.7\% & 41.3\% & 65.6\% & 58.6\% \\ \hline
AdaAttN & 130 & 0.125 & 0.304 & 50.8\% & 38.9\% & \underline{63.0}\% & \underline{58.3}\% \\ \hline
MCCNet & 29 & 0.137 & 0.308 & 45.3\% & 36.2\% & 73.6\% & 70.8\% \\ \hline
ArtFlow & 168 & \underline{0.121} & 0.314 & 44.2\% & 39.4\% & 58.8\% & 55.5\% \\ \hline
AdaIN & \underline{11} & 0.160 & 0.336 & 51.0\% & 27.8\% & 72.4\% & 64.6\% \\ \hline \hline
UCAST+AdaIN & \underline{11} & \textbf{0.117} & \underline{0.302} & \underline{64.2\%} & - & \textbf{39.2}\% & \textbf{36.3}\% \\ \hline
UCAST+StyleTr$^2$ & 87 & 0.122 & 0.311 & \textbf{68.2}\% & - & - & - \\ \hline
UCAST+ArtFlow & 168 & \underline{0.121} & \textbf{0.251} & 62.0\% & - & - & - \\
\bottomrule
\end{tabular}
%}%resizebox
\label{tab:quantitative}
\end{table*}

%%%% Tables/tab_quantitative.tex ends here %%%%

%%%% Figs/fig_backbone.tex starts here %%%%

\begin{figure*}
\newcommand\backbonefigurewidth{0.245}
\centering
\includegraphics[width=0.99\linewidth]{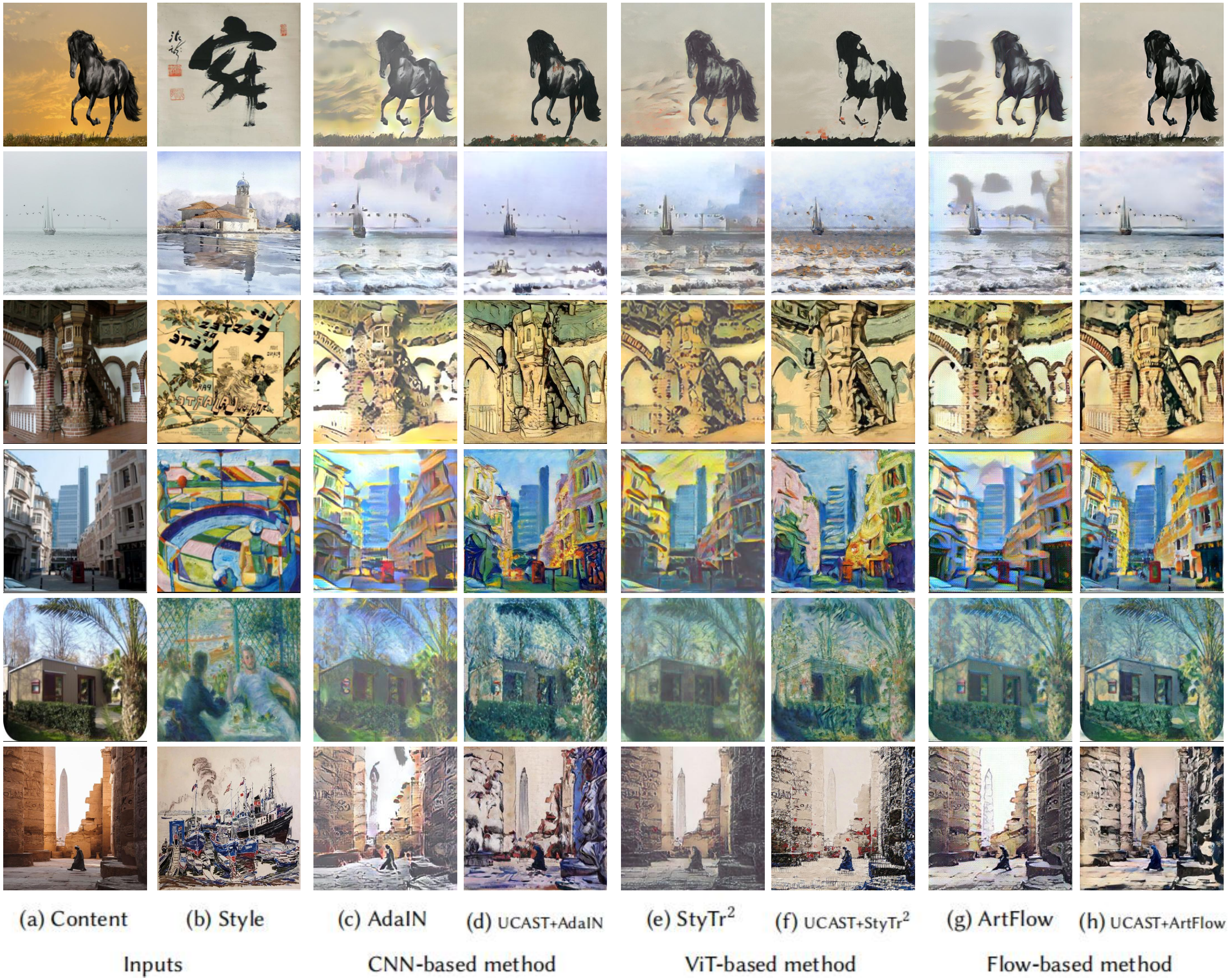}
\caption{Qualitative comparisons on different backbones trained under our UCAST framework.
}
\label{fig:backbone}
\end{figure*}

%%%% Figs/fig_backbone.tex ends here %%%%

%%%% 4-Experiments.tex starts here %%%%
\section{Experiments}

We compare UCAST with several state-of-the-art style transfer methods, including AdaIN~\cite{Huang:2017:AdaIn}, ArtFlow~\cite{An:2021:Artflow}, MCCNet~\cite{Deng:2021:MCC},  AdaAttN~\cite{liu2021adaattn}, IEST~\cite{chen2021artistic}, StyleFormer~\cite{Wu:2021:SF}, as well as StyTr$^2$~\cite{deng2021stytr2}.
All the baselines are trained using publicly available implementations with default configurations.
The comparison of inference speed is shown in Table~\ref{tab:quantitative}.
In all our experiments, our results are generated by using AdaIN as backbone, if there is no specific annotation.

\paragraph{Implementation details}
We collect 100,000 artistic images in different styles from WikiArt~\cite{Phillips:2011:wikiart} and randomly sample 20,000 images as our artistic dataset.
We averagely sample 20,000 images from Places365~\cite{Zhou:2018:Places365} as realistic image dataset.
We train and evaluate our framework on those artistic and realistic images.
In the training phase, all images are loaded with $256 \times 256$ resolution.
The number of feature map layers $M$ is set to be 4.
The dimension $K$ of style latent code is set to 512, 512, 512, and 512 for the four different layers, respectively.
We use Adam~\cite{kingma2014adam} as optimizer with $\beta_1=0.5$, $\beta_2=0.999$, and a batch size of $4$.
The initial learning rate is set to $1 \times 10^{-4}$ and linear decayed linear for total $8 \times 10^5$ iterations.
The training process takes about $18$ hours on one NVIDIA GeForce RTX3090.

\subsection{Effectiveness on Various Backbones}
Our UCAST, as an separated network structure can be plug-and-play for most arbitrary image style transfer models.
In our experiments, we adopt UCAST to AdaIN~\cite{Huang:2017:AdaIn}, ArtFlow~\cite{An:2021:Artflow}, and StyTr$^2$~\cite{deng2021stytr2}.
AdaIN~\cite{Huang:2017:AdaIn} is a CNN-based style transfer model that includes a fixed VGG network to encode the content and style images, an adaptive instance normalization layer to align the channel-wise mean and variance of content features to match those of style features, and a CNN decoder to invert the AdaIN output to the image spaces.
ArtFlow~\cite{An:2021:Artflow} is a neural flow-based model which consists of reversible neural flows and an unbiased feature transfer module.
Neural flows are a type of deep generative model that learns the precise likelihood of high-dimensional observations via a series of invertible transformations.
StyTr$^2$~\cite{deng2021stytr2} is a ViT-based model that contains two transformer encoders for the content image and the style reference respectively, a multi-layer transformer decoder for content sequence stylization, and a CNN decoder.

The comparison results are shown in Fig.~\ref{fig:backbone}.
When transferring style image of ink and wash, as shown in the 1\textsuperscript{st} row, the three backbone methods cannot faithfully generate the brush strokes and the empty background.
By training under the UCAST framework, all the enhanced methods can generate high quality ink and wash images with smooth empty background and vivid strokes.
When dealing with watercolor image, as shown in the 2\textsuperscript{nd} row, the backbones cannot capture the feeling of color blooming.
Since the sky in the content image is a large empty area which the style image does not have, the three backbones tend to generate obvious artifacts.
Being trained under UCAST can reduce the artifacts obviously and transfer the unique strokes of watercolor.
As shown in the 3\textsuperscript{rd} and 4\textsuperscript{th} rows, the backbones fail to transfer the sharp lines in the style reference, while UCAST improves the details of the generated images significantly.
UCAST can also help all the backbones to generate vivid brush strokes of oil paintings, as shown in the 5\textsuperscript{th} row.

%%%% Figs/fig_sota_compare.tex starts here %%%%

\begin{figure*}[htp]
\newcommand{\galleryfigurewidth}{0.092}
\centering
%%%%
    \begin{minipage}[t]{\textwidth}
    \centering
        \begin{minipage}{\galleryfigurewidth\linewidth}
        \includegraphics[width=\linewidth]{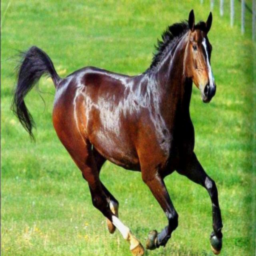}
        \end{minipage}
        \begin{minipage}{\galleryfigurewidth\linewidth}
        \includegraphics[width=\linewidth]{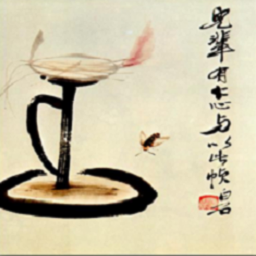}
        \end{minipage}
        \begin{minipage}{\galleryfigurewidth\linewidth}
        \includegraphics[width=\linewidth]{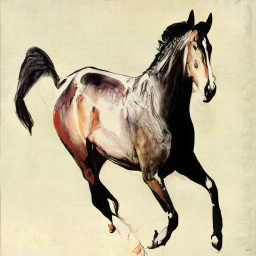}
        \end{minipage}
        \begin{minipage}{\galleryfigurewidth\linewidth}
        \includegraphics[width=\linewidth]{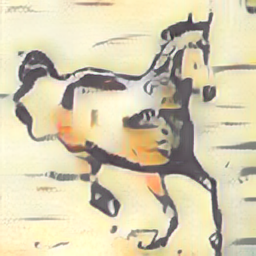}
        \end{minipage}
        \begin{minipage}{\galleryfigurewidth\linewidth}
        \includegraphics[width=\linewidth]{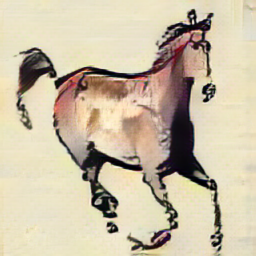}
        \end{minipage}
        \begin{minipage}{\galleryfigurewidth\linewidth}
        \includegraphics[width=\linewidth]{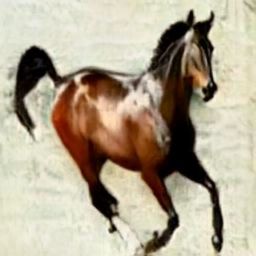}
        \end{minipage}
        \begin{minipage}{\galleryfigurewidth\linewidth}
        \includegraphics[width=\linewidth]{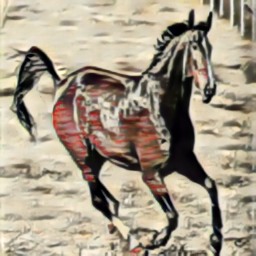}
        \end{minipage}
        \begin{minipage}{\galleryfigurewidth\linewidth}
        \includegraphics[width=\linewidth]{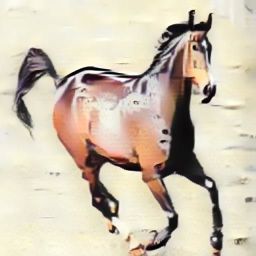}
        \end{minipage}
        \begin{minipage}{\galleryfigurewidth\linewidth}
        \includegraphics[width=\linewidth]{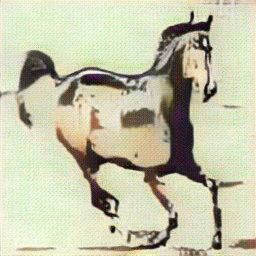}
        \end{minipage}
        \begin{minipage}{\galleryfigurewidth\linewidth}
        \includegraphics[width=\linewidth]{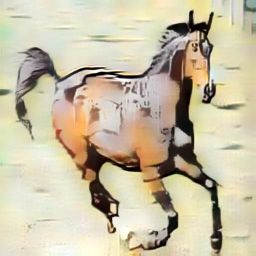}
        \end{minipage}
    \end{minipage}
    \begin{minipage}[t]{\textwidth}
    \centering
        \begin{minipage}{\galleryfigurewidth\linewidth}
        \includegraphics[width=\linewidth]{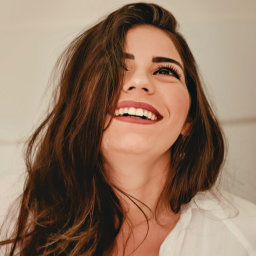}
        \end{minipage}
        \begin{minipage}{\galleryfigurewidth\linewidth}
        \includegraphics[width=\linewidth]{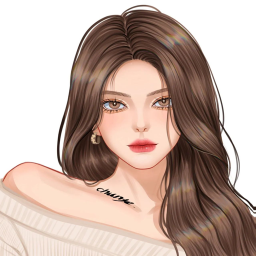}
        \end{minipage}
        \begin{minipage}{\galleryfigurewidth\linewidth}
        \includegraphics[width=\linewidth]{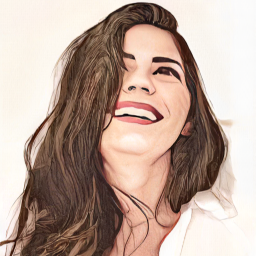}
        \end{minipage}
        \begin{minipage}{\galleryfigurewidth\linewidth}
        \includegraphics[width=\linewidth]{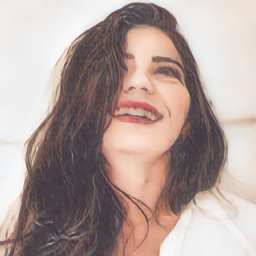}
        \end{minipage}
        \begin{minipage}{\galleryfigurewidth\linewidth}
        \includegraphics[width=\linewidth]{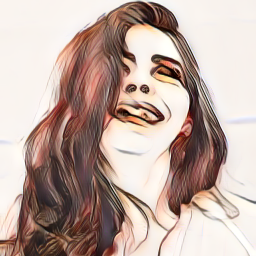}
        \end{minipage}
        \begin{minipage}{\galleryfigurewidth\linewidth}
        \includegraphics[width=\linewidth]{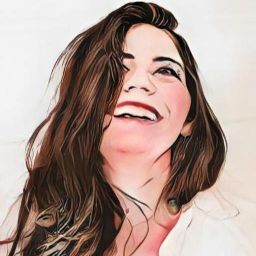}
        \end{minipage}
        \begin{minipage}{\galleryfigurewidth\linewidth}
        \includegraphics[width=\linewidth]{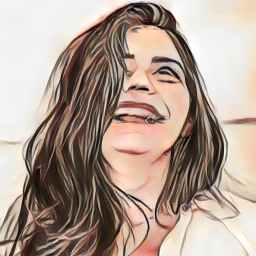}
        \end{minipage}
        \begin{minipage}{\galleryfigurewidth\linewidth}
        \includegraphics[width=\linewidth]{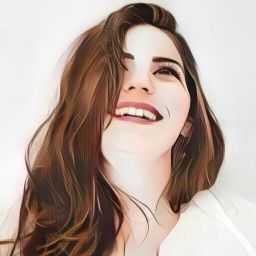}
        \end{minipage}
        \begin{minipage}{\galleryfigurewidth\linewidth}
        \includegraphics[width=\linewidth]{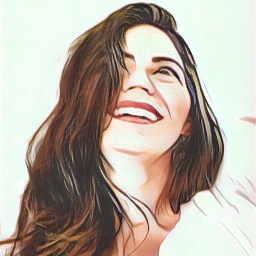}
        \end{minipage}
        \begin{minipage}{\galleryfigurewidth\linewidth}
        \includegraphics[width=\linewidth]{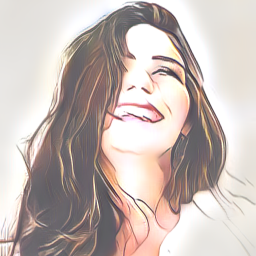}
        \end{minipage}
    \end{minipage}
    \begin{minipage}[t]{\textwidth}
    \centering
        \begin{minipage}{\galleryfigurewidth\linewidth}
        \includegraphics[width=\linewidth]{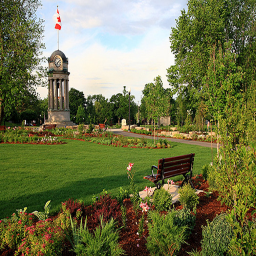}
        \end{minipage}
        \begin{minipage}{\galleryfigurewidth\linewidth}
        \includegraphics[width=\linewidth]{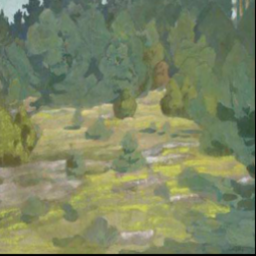}
        \end{minipage}
        \begin{minipage}{\galleryfigurewidth\linewidth}
        \includegraphics[width=\linewidth]{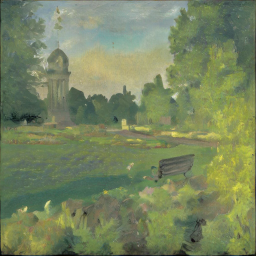}
        \end{minipage}
        \begin{minipage}{\galleryfigurewidth\linewidth}
        \includegraphics[width=\linewidth]{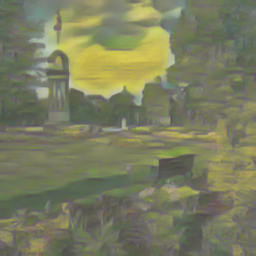}
        \end{minipage}
        \begin{minipage}{\galleryfigurewidth\linewidth}
        \includegraphics[width=\linewidth]{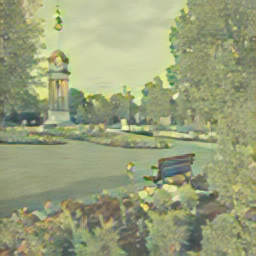}
        \end{minipage}
        \begin{minipage}{\galleryfigurewidth\linewidth}
        \includegraphics[width=\linewidth]{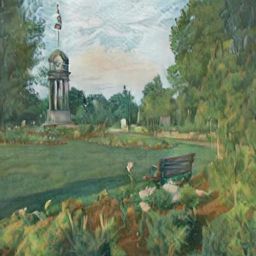}
        \end{minipage}
        \begin{minipage}{\galleryfigurewidth\linewidth}
        \includegraphics[width=\linewidth]{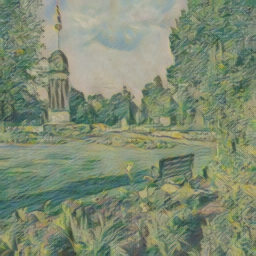}
        \end{minipage}
        \begin{minipage}{\galleryfigurewidth\linewidth}
        \includegraphics[width=\linewidth]{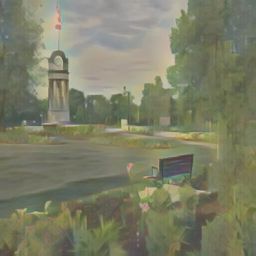}
        \end{minipage}
        \begin{minipage}{\galleryfigurewidth\linewidth}
        \includegraphics[width=\linewidth]{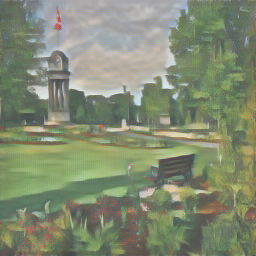}
        \end{minipage}
        \begin{minipage}{\galleryfigurewidth\linewidth}
        \includegraphics[width=\linewidth]{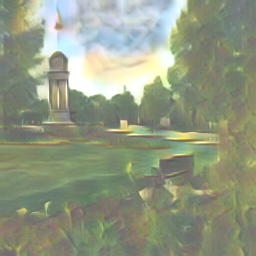}
        \end{minipage}
    \end{minipage}
    \begin{minipage}[t]{\textwidth}
    \centering
        \begin{minipage}{\galleryfigurewidth\linewidth}
        \includegraphics[width=\linewidth]{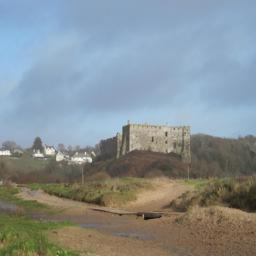}
        \end{minipage}
        \begin{minipage}{\galleryfigurewidth\linewidth}
        \includegraphics[width=\linewidth]{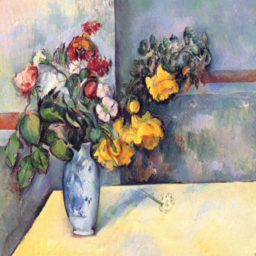}
        \end{minipage}
        \begin{minipage}{\galleryfigurewidth\linewidth}
        \includegraphics[width=\linewidth]{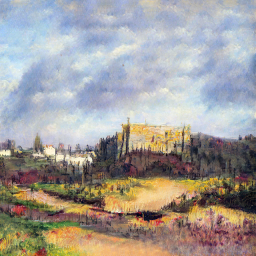}
        \end{minipage}
        \begin{minipage}{\galleryfigurewidth\linewidth}
        \includegraphics[width=\linewidth]{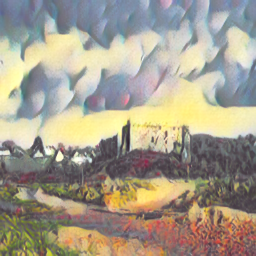}
        \end{minipage}
        \begin{minipage}{\galleryfigurewidth\linewidth}
        \includegraphics[width=\linewidth]{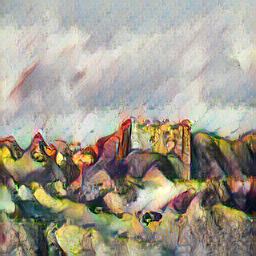}
        \end{minipage}
        \begin{minipage}{\galleryfigurewidth\linewidth}
        \includegraphics[width=\linewidth]{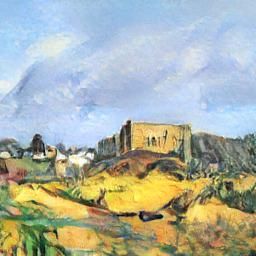}
        \end{minipage}
        \begin{minipage}{\galleryfigurewidth\linewidth}
        \includegraphics[width=\linewidth]{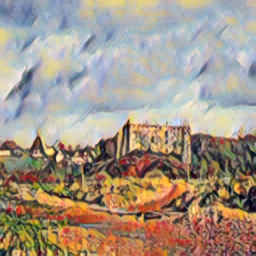}
        \end{minipage}
        \begin{minipage}{\galleryfigurewidth\linewidth}
        \includegraphics[width=\linewidth]{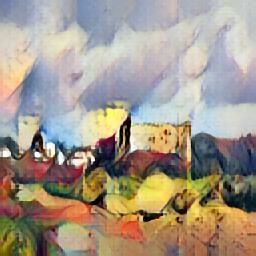}
        \end{minipage}
        \begin{minipage}{\galleryfigurewidth\linewidth}
        \includegraphics[width=\linewidth]{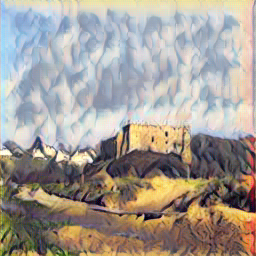}
        \end{minipage}
        \begin{minipage}{\galleryfigurewidth\linewidth}
        \includegraphics[width=\linewidth]{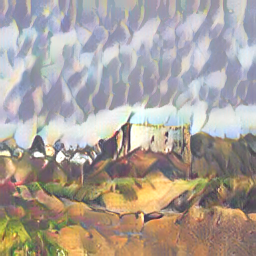}
        \end{minipage}
    \end{minipage}
    \begin{minipage}[t]{\textwidth}
    \centering
        \begin{minipage}{\galleryfigurewidth\linewidth}
        \includegraphics[width=\linewidth]{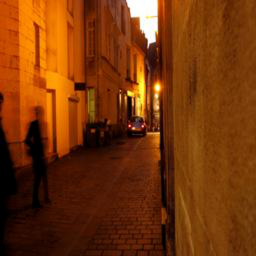}
        \end{minipage}
        \begin{minipage}{\galleryfigurewidth\linewidth}
        \includegraphics[width=\linewidth]{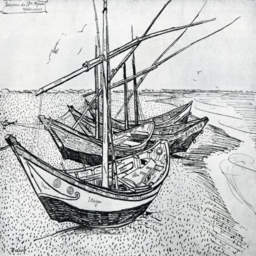}
        \end{minipage}
        \begin{minipage}{\galleryfigurewidth\linewidth}
        \includegraphics[width=\linewidth]{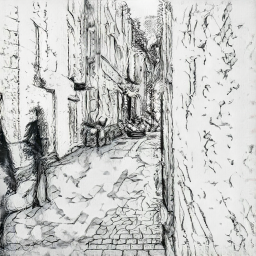}
        \end{minipage}
        \begin{minipage}{\galleryfigurewidth\linewidth}
        \includegraphics[width=\linewidth]{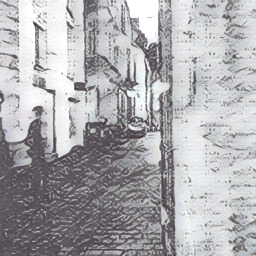}
        \end{minipage}
        \begin{minipage}{\galleryfigurewidth\linewidth}
        \includegraphics[width=\linewidth]{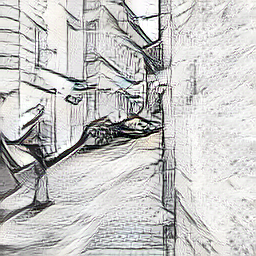}
        \end{minipage}
        \begin{minipage}{\galleryfigurewidth\linewidth}
        \includegraphics[width=\linewidth]{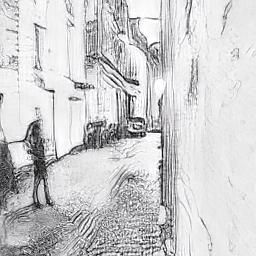}
        \end{minipage}
        \begin{minipage}{\galleryfigurewidth\linewidth}
        \includegraphics[width=\linewidth]{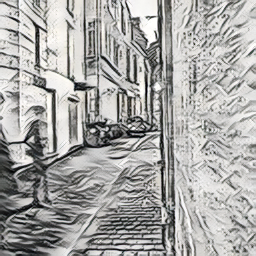}
        \end{minipage}
        \begin{minipage}{\galleryfigurewidth\linewidth}
        \includegraphics[width=\linewidth]{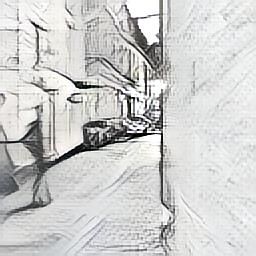}
        \end{minipage}
        \begin{minipage}{\galleryfigurewidth\linewidth}
        \includegraphics[width=\linewidth]{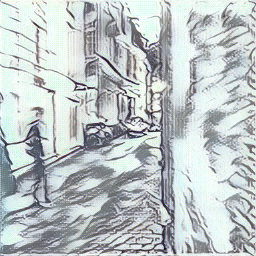}
        \end{minipage}
        \begin{minipage}{\galleryfigurewidth\linewidth}
        \includegraphics[width=\linewidth]{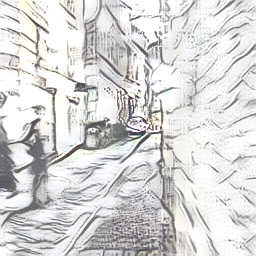}
        \end{minipage}
    \end{minipage}
    \begin{minipage}[t]{\textwidth}
    \centering
        \begin{minipage}{\galleryfigurewidth\linewidth}
        \includegraphics[width=\linewidth]{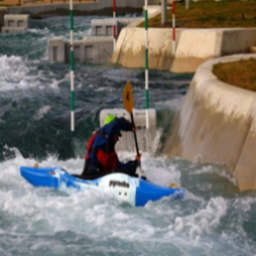}
        \end{minipage}
        \begin{minipage}{\galleryfigurewidth\linewidth}
        \includegraphics[width=\linewidth]{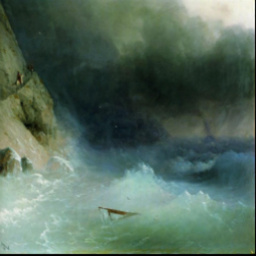}
        \end{minipage}
        \begin{minipage}{\galleryfigurewidth\linewidth}
        \includegraphics[width=\linewidth]{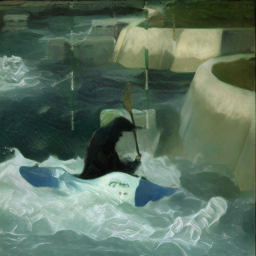}
        \end{minipage}
        \begin{minipage}{\galleryfigurewidth\linewidth}
        \includegraphics[width=\linewidth]{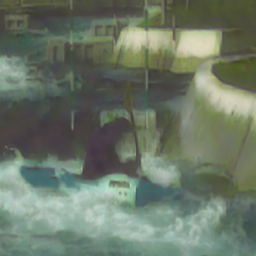}
        \end{minipage}
        \begin{minipage}{\galleryfigurewidth\linewidth}
        \includegraphics[width=\linewidth]{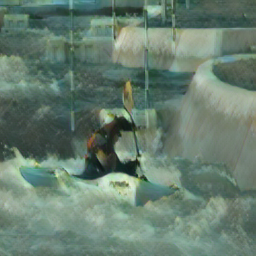}
        \end{minipage}
        \begin{minipage}{\galleryfigurewidth\linewidth}
        \includegraphics[width=\linewidth]{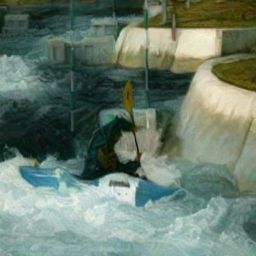}
        \end{minipage}
        \begin{minipage}{\galleryfigurewidth\linewidth}
        \includegraphics[width=\linewidth]{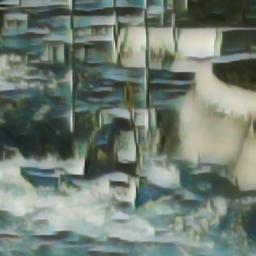}
        \end{minipage}
        \begin{minipage}{\galleryfigurewidth\linewidth}
        \includegraphics[width=\linewidth]{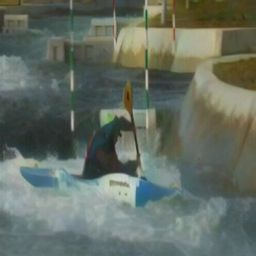}
        \end{minipage}
        \begin{minipage}{\galleryfigurewidth\linewidth}
        \includegraphics[width=\linewidth]{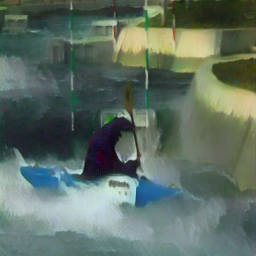}
        \end{minipage}
        \begin{minipage}{\galleryfigurewidth\linewidth}
        \includegraphics[width=\linewidth]{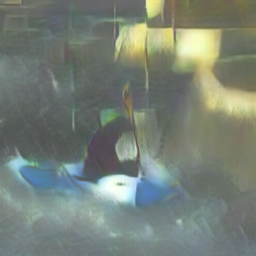}
        \end{minipage}
    \end{minipage}
    \begin{minipage}[t]{\textwidth}
    \centering
        \begin{minipage}{\galleryfigurewidth\linewidth}
        \includegraphics[width=\linewidth]{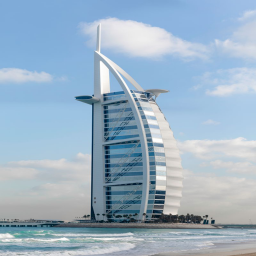}
        \end{minipage}
        \begin{minipage}{\galleryfigurewidth\linewidth}
        \includegraphics[width=\linewidth]{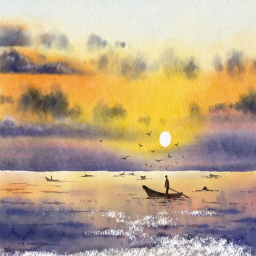}
        \end{minipage}
        \begin{minipage}{\galleryfigurewidth\linewidth}
        \includegraphics[width=\linewidth]{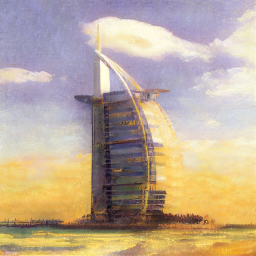}
        \end{minipage}
        \begin{minipage}{\galleryfigurewidth\linewidth}
        \includegraphics[width=\linewidth]{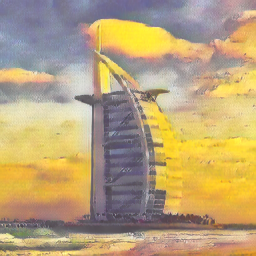}
        \end{minipage}
        \begin{minipage}{\galleryfigurewidth\linewidth}
        \includegraphics[width=\linewidth]{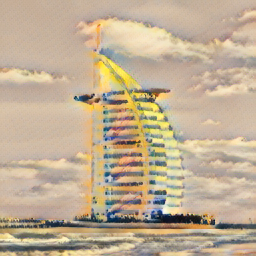}
        \end{minipage}
        \begin{minipage}{\galleryfigurewidth\linewidth}
        \includegraphics[width=\linewidth]{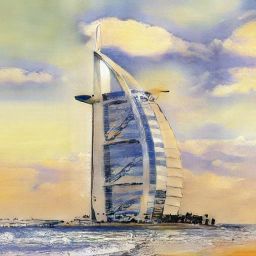}
        \end{minipage}
        \begin{minipage}{\galleryfigurewidth\linewidth}
        \includegraphics[width=\linewidth]{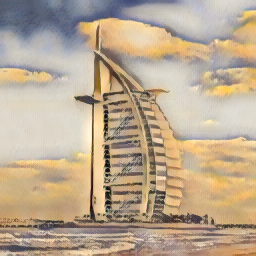}
        \end{minipage}
        \begin{minipage}{\galleryfigurewidth\linewidth}
        \includegraphics[width=\linewidth]{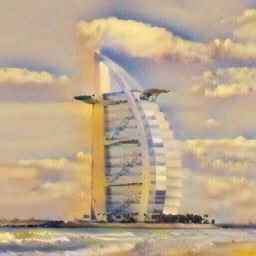}
        \end{minipage}
        \begin{minipage}{\galleryfigurewidth\linewidth}
        \includegraphics[width=\linewidth]{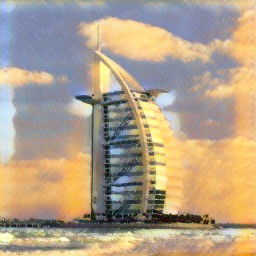}
        \end{minipage}
        \begin{minipage}{\galleryfigurewidth\linewidth}
        \includegraphics[width=\linewidth]{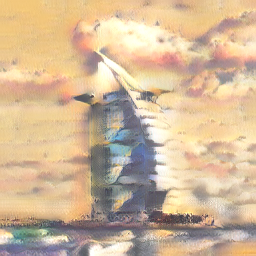}
        \end{minipage}
    \end{minipage}
    % \begin{minipage}[t]{\textwidth}
    % \centering
    %     \begin{minipage}{\galleryfigurewidth\linewidth}
    %     \includegraphics[width=\linewidth]{Image/sota/e/e_content}
    %     \end{minipage}
    %     \begin{minipage}{\galleryfigurewidth\linewidth}
    %     \includegraphics[width=\linewidth]{Image/sota/e/e_style}
    %     \end{minipage}
    %     \begin{minipage}{\galleryfigurewidth\linewidth}
    %     \includegraphics[width=\linewidth]{Image/sota/e/e_UCAST}
    %     \end{minipage}
    %     \begin{minipage}{\galleryfigurewidth\linewidth}
    %     \includegraphics[width=\linewidth]{Image/sota/e/e_stytr2}
    %     \end{minipage}
    %     \begin{minipage}{\galleryfigurewidth\linewidth}
    %     \includegraphics[width=\linewidth]{Image/sota/e/e_styleformer}
    %     \end{minipage}
    %     \begin{minipage}{\galleryfigurewidth\linewidth}
    %     \includegraphics[width=\linewidth]{Image/sota/e/e_IEST}
    %     \end{minipage}
    %     \begin{minipage}{\galleryfigurewidth\linewidth}
    %     \includegraphics[width=\linewidth]{Image/sota/e/e_adaattn}
    %     \end{minipage}
    %     \begin{minipage}{\galleryfigurewidth\linewidth}
    %     \includegraphics[width=\linewidth]{Image/sota/e/e_mcc}
    %     \end{minipage}
    %     \begin{minipage}{\galleryfigurewidth\linewidth}
    %     \includegraphics[width=\linewidth]{Image/sota/e/e_artflow}
    %     \end{minipage}
    %     \begin{minipage}{\galleryfigurewidth\linewidth}
    %     \includegraphics[width=\linewidth]{Image/sota/e/e_adain}
    %     \end{minipage}
    % \end{minipage}
    \begin{minipage}[t]{\textwidth}
    \centering
        \begin{minipage}{\galleryfigurewidth\linewidth}
        \includegraphics[width=\linewidth]{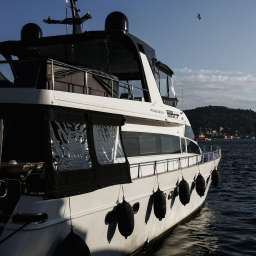}
        \end{minipage}
        \begin{minipage}{\galleryfigurewidth\linewidth}
        \includegraphics[width=\linewidth]{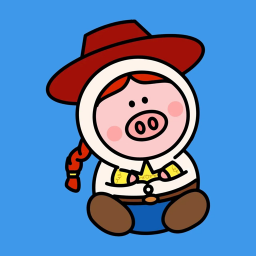}
        \end{minipage}
        \begin{minipage}{\galleryfigurewidth\linewidth}
        \includegraphics[width=\linewidth]{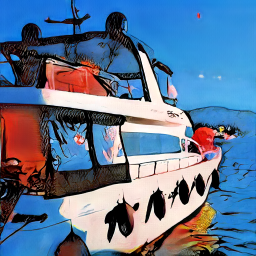}
        \end{minipage}
        \begin{minipage}{\galleryfigurewidth\linewidth}
        \includegraphics[width=\linewidth]{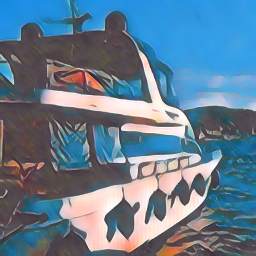}
        \end{minipage}
        \begin{minipage}{\galleryfigurewidth\linewidth}
        \includegraphics[width=\linewidth]{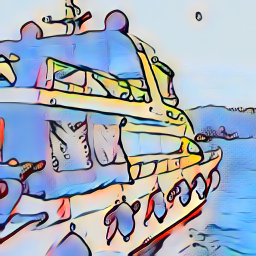}
        \end{minipage}
        \begin{minipage}{\galleryfigurewidth\linewidth}
        \includegraphics[width=\linewidth]{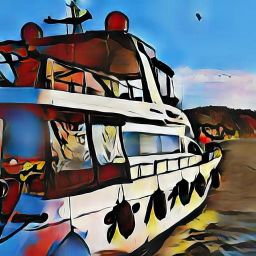}
        \end{minipage}
        \begin{minipage}{\galleryfigurewidth\linewidth}
        \includegraphics[width=\linewidth]{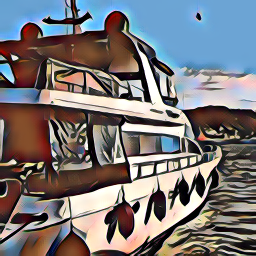}
        \end{minipage}
        \begin{minipage}{\galleryfigurewidth\linewidth}
        \includegraphics[width=\linewidth]{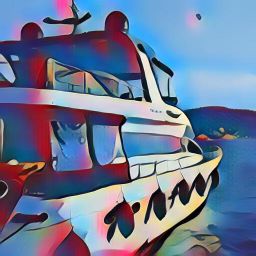}
        \end{minipage}
        \begin{minipage}{\galleryfigurewidth\linewidth}
        \includegraphics[width=\linewidth]{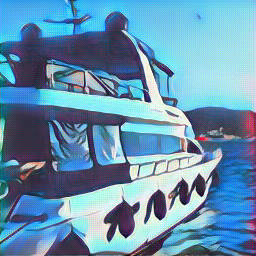}
        \end{minipage}
        \begin{minipage}{\galleryfigurewidth\linewidth}
        \includegraphics[width=\linewidth]{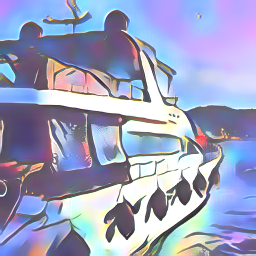}
        \end{minipage}
    \end{minipage}
    \begin{minipage}[t]{\textwidth}
    \centering
        \begin{minipage}{\galleryfigurewidth\linewidth}
        \includegraphics[width=\linewidth]{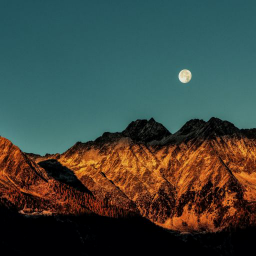}
        \end{minipage}
        \begin{minipage}{\galleryfigurewidth\linewidth}
        \includegraphics[width=\linewidth]{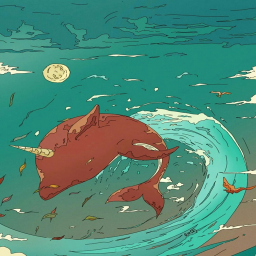}
        \end{minipage}
        \begin{minipage}{\galleryfigurewidth\linewidth}
        \includegraphics[width=\linewidth]{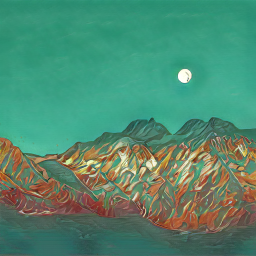}
        \end{minipage}
        \begin{minipage}{\galleryfigurewidth\linewidth}
        \includegraphics[width=\linewidth]{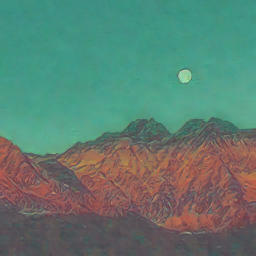}
        \end{minipage}
        \begin{minipage}{\galleryfigurewidth\linewidth}
        \includegraphics[width=\linewidth]{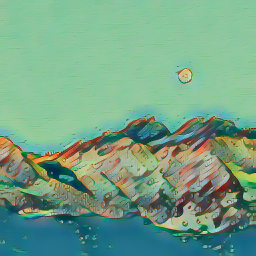}
        \end{minipage}
        \begin{minipage}{\galleryfigurewidth\linewidth}
        \includegraphics[width=\linewidth]{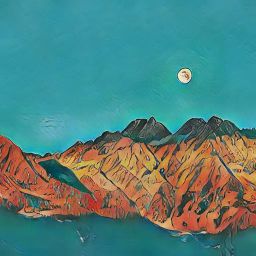}
        \end{minipage}
        \begin{minipage}{\galleryfigurewidth\linewidth}
        \includegraphics[width=\linewidth]{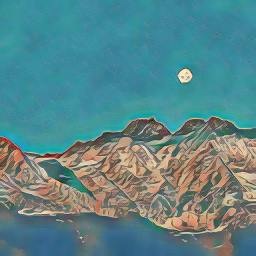}
        \end{minipage}
        \begin{minipage}{\galleryfigurewidth\linewidth}
        \includegraphics[width=\linewidth]{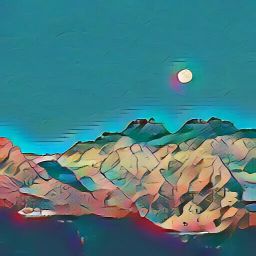}
        \end{minipage}
        \begin{minipage}{\galleryfigurewidth\linewidth}
        \includegraphics[width=\linewidth]{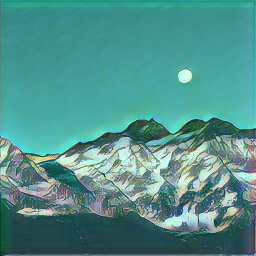}
        \end{minipage}
        \begin{minipage}{\galleryfigurewidth\linewidth}
        \includegraphics[width=\linewidth]{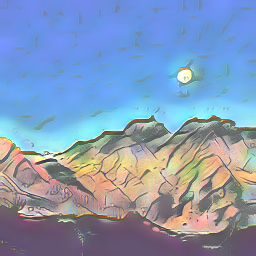}
        \end{minipage}
    \end{minipage}
    \begin{minipage}[t]{\textwidth}
    \centering
        \begin{minipage}{\galleryfigurewidth\linewidth}
        \includegraphics[width=\linewidth]{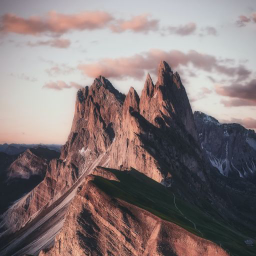}
        \end{minipage}
        \begin{minipage}{\galleryfigurewidth\linewidth}
        \includegraphics[width=\linewidth]{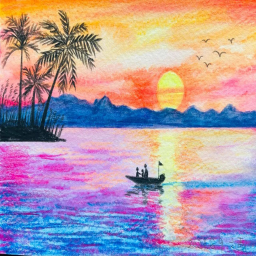}
        \end{minipage}
        \begin{minipage}{\galleryfigurewidth\linewidth}
        \includegraphics[width=\linewidth]{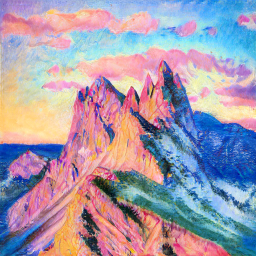}
        \end{minipage}
        \begin{minipage}{\galleryfigurewidth\linewidth}
        \includegraphics[width=\linewidth]{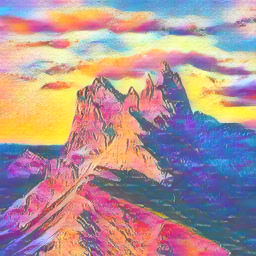}
        \end{minipage}
        \begin{minipage}{\galleryfigurewidth\linewidth}
        \includegraphics[width=\linewidth]{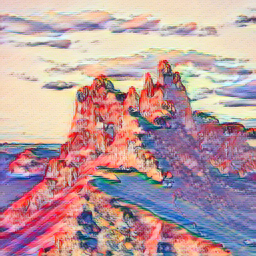}
        \end{minipage}
        \begin{minipage}{\galleryfigurewidth\linewidth}
        \includegraphics[width=\linewidth]{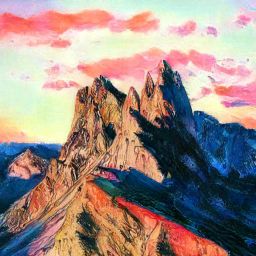}
        \end{minipage}
        \begin{minipage}{\galleryfigurewidth\linewidth}
        \includegraphics[width=\linewidth]{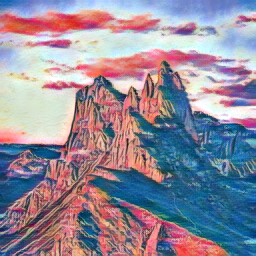}
        \end{minipage}
        \begin{minipage}{\galleryfigurewidth\linewidth}
        \includegraphics[width=\linewidth]{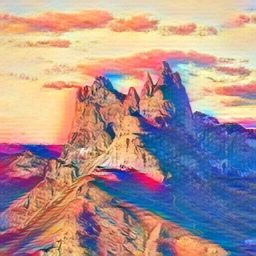}
        \end{minipage}
        \begin{minipage}{\galleryfigurewidth\linewidth}
        \includegraphics[width=\linewidth]{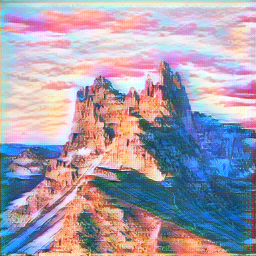}
        \end{minipage}
        \begin{minipage}{\galleryfigurewidth\linewidth}
        \includegraphics[width=\linewidth]{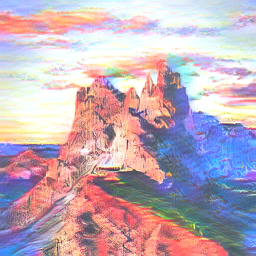}
        \end{minipage}
    \end{minipage}
    \begin{minipage}[t]{\textwidth}
    \centering
        \begin{minipage}{\galleryfigurewidth\linewidth}
        \includegraphics[width=\linewidth]{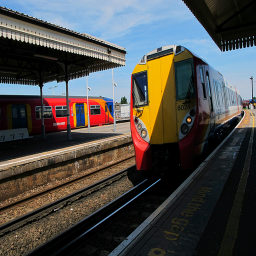}
        \end{minipage}
        \begin{minipage}{\galleryfigurewidth\linewidth}
        \includegraphics[width=\linewidth]{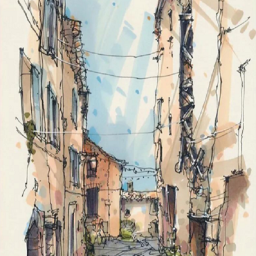}
        \end{minipage}
        \begin{minipage}{\galleryfigurewidth\linewidth}
        \includegraphics[width=\linewidth]{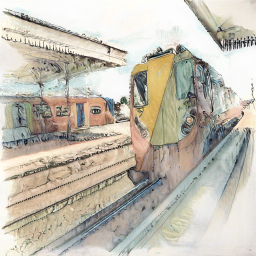}
        \end{minipage}
        \begin{minipage}{\galleryfigurewidth\linewidth}
        \includegraphics[width=\linewidth]{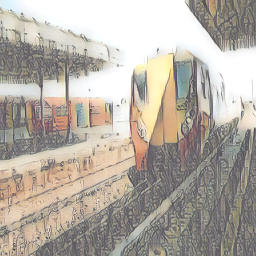}
        \end{minipage}
        \begin{minipage}{\galleryfigurewidth\linewidth}
        \includegraphics[width=\linewidth]{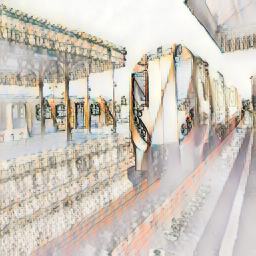}
        \end{minipage}
        \begin{minipage}{\galleryfigurewidth\linewidth}
        \includegraphics[width=\linewidth]{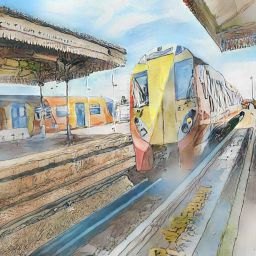}
        \end{minipage}
        \begin{minipage}{\galleryfigurewidth\linewidth}
        \includegraphics[width=\linewidth]{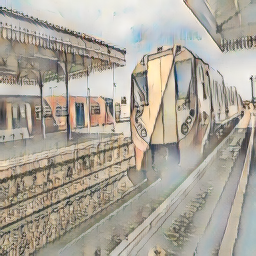}
        \end{minipage}
        \begin{minipage}{\galleryfigurewidth\linewidth}
        \includegraphics[width=\linewidth]{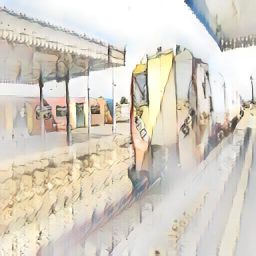}
        \end{minipage}
        \begin{minipage}{\galleryfigurewidth\linewidth}
        \includegraphics[width=\linewidth]{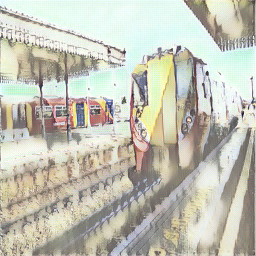}
        \end{minipage}
        \begin{minipage}{\galleryfigurewidth\linewidth}
        \includegraphics[width=\linewidth]{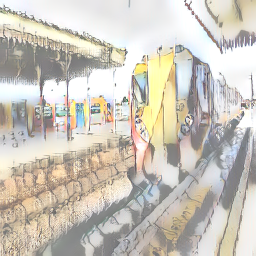}
        \end{minipage}
    \end{minipage}
    \begin{minipage}[t]{\textwidth}
    \centering
        \begin{minipage}{\galleryfigurewidth\linewidth}
        \includegraphics[width=\linewidth]{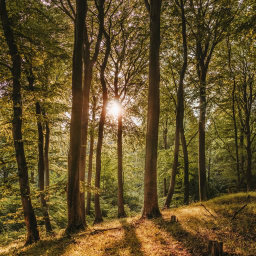}
        \end{minipage}
        \begin{minipage}{\galleryfigurewidth\linewidth}
        \includegraphics[width=\linewidth]{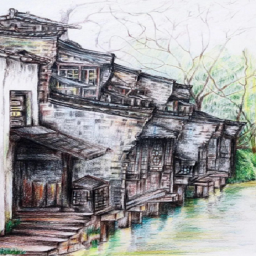}
        \end{minipage}
        \begin{minipage}{\galleryfigurewidth\linewidth}
        \includegraphics[width=\linewidth]{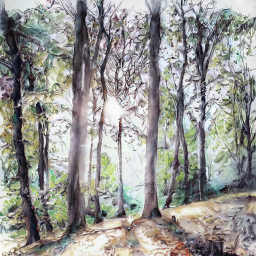}
        \end{minipage}
        \begin{minipage}{\galleryfigurewidth\linewidth}
        \includegraphics[width=\linewidth]{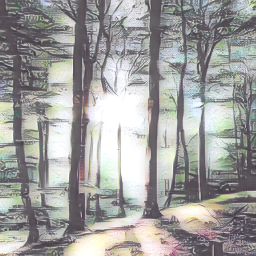}
        \end{minipage}
        \begin{minipage}{\galleryfigurewidth\linewidth}
        \includegraphics[width=\linewidth]{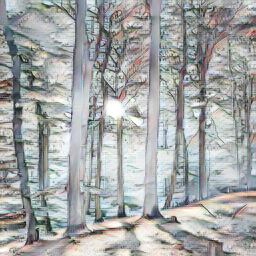}
        \end{minipage}
        \begin{minipage}{\galleryfigurewidth\linewidth}
        \includegraphics[width=\linewidth]{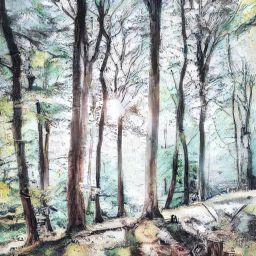}
        \end{minipage}
        \begin{minipage}{\galleryfigurewidth\linewidth}
        \includegraphics[width=\linewidth]{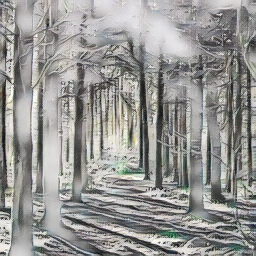}
        \end{minipage}
        \begin{minipage}{\galleryfigurewidth\linewidth}
        \includegraphics[width=\linewidth]{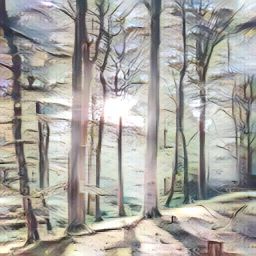}
        \end{minipage}
        \begin{minipage}{\galleryfigurewidth\linewidth}
        \includegraphics[width=\linewidth]{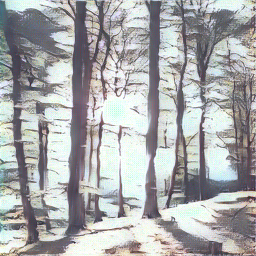}
        \end{minipage}
        \begin{minipage}{\galleryfigurewidth\linewidth}
        \includegraphics[width=\linewidth]{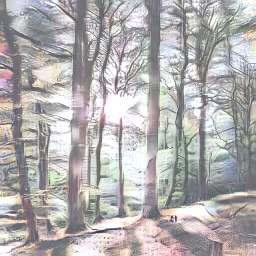}
        \end{minipage}
    \end{minipage}
\begin{subfigure}[t]{\galleryfigurewidth\linewidth}
    \subcaption*{Content}
    \label{fig:SOTA_a}
\end{subfigure}
\begin{subfigure}[t]{\galleryfigurewidth\linewidth}
    \subcaption*{Style}
    \label{fig:SOTA_b}
\end{subfigure}
\begin{subfigure}[t]{\galleryfigurewidth\linewidth}
    \subcaption*{Ours}
    \label{fig:SOTA_c}
\end{subfigure}
\begin{subfigure}[t]{\galleryfigurewidth\linewidth}
    \subcaption*{StyTr$^2$}
    \label{fig:SOTA_d}
\end{subfigure}
\begin{subfigure}[t]{\galleryfigurewidth\linewidth}
    \subcaption*{StyleFormer}
    \label{fig:SOTA_e}
\end{subfigure}
\begin{subfigure}[t]{\galleryfigurewidth\linewidth}
 \subcaption*{IEST}
    \label{fig:SOTA_f}
\end{subfigure}
\begin{subfigure}[t]{\galleryfigurewidth\linewidth}
    \subcaption*{AdaAttN}
    \label{fig:SOTA_g}
\end{subfigure}
\begin{subfigure}[t]{\galleryfigurewidth\linewidth}
    \subcaption*{MCCNet}
    \label{fig:SOTA_h}
\end{subfigure}
\begin{subfigure}[t]{\galleryfigurewidth\linewidth}
    \subcaption*{ArtFlow}
    \label{fig:SOTA_i}
\end{subfigure}
\begin{subfigure}[t]{\galleryfigurewidth\linewidth}
    \subcaption*{AdaIN}
    \label{fig:SOTA_j}
\end{subfigure}

\caption{
Qualitative comparisons with several state-of-the-art style transfer methods, including StyTr$^2$~\cite{deng2021stytr2}, StyleFormer~\cite{Wu:2021:SF}, IEST~\cite{chen2021artistic}, AdaAttN~\cite{liu2021adaattn}, 
MCCNet~\cite{Deng:2021:MCC}, ArtFlow~\cite{An:2021:Artflow}, 
AdaIN~\cite{Huang:2017:AdaIn}.
% Qualitative comparisons with several state-of-the-art style transfer methods, including IEST~\cite{chen2021artistic}, AdaAttN~\cite{liu2021adaattn}, MCCNet~\cite{Deng:2021:MCC}, ArtFlow~\cite{An:2021:Artflow}, SANet~\cite{Park:2019:AST}, LST~\cite{Li:2019:LLT}, AdaIN~\cite{Huang:2017:AdaIn}, and NST~\cite{Gatys:2016:IST}.
% Content image credits: \emph{Horse} Pixabay/Pexels (CC0), \emph{Cat} Pixabay/Pexels (CC0).
% Style image credits (from the the 2\textsuperscript{nd} row to the 7\textsuperscript{th} row): \{Richard Parkes Bonington, Philip William May, Michel Ange Corneille, Claude Monet, Vincent van Gogh, Childe Hassam\}/The Art Institute of Chicago (CC0).
}
\label{fig:SOTA}
\end{figure*}

\begin{figure}
\centering
\includegraphics[width=0.99\linewidth]{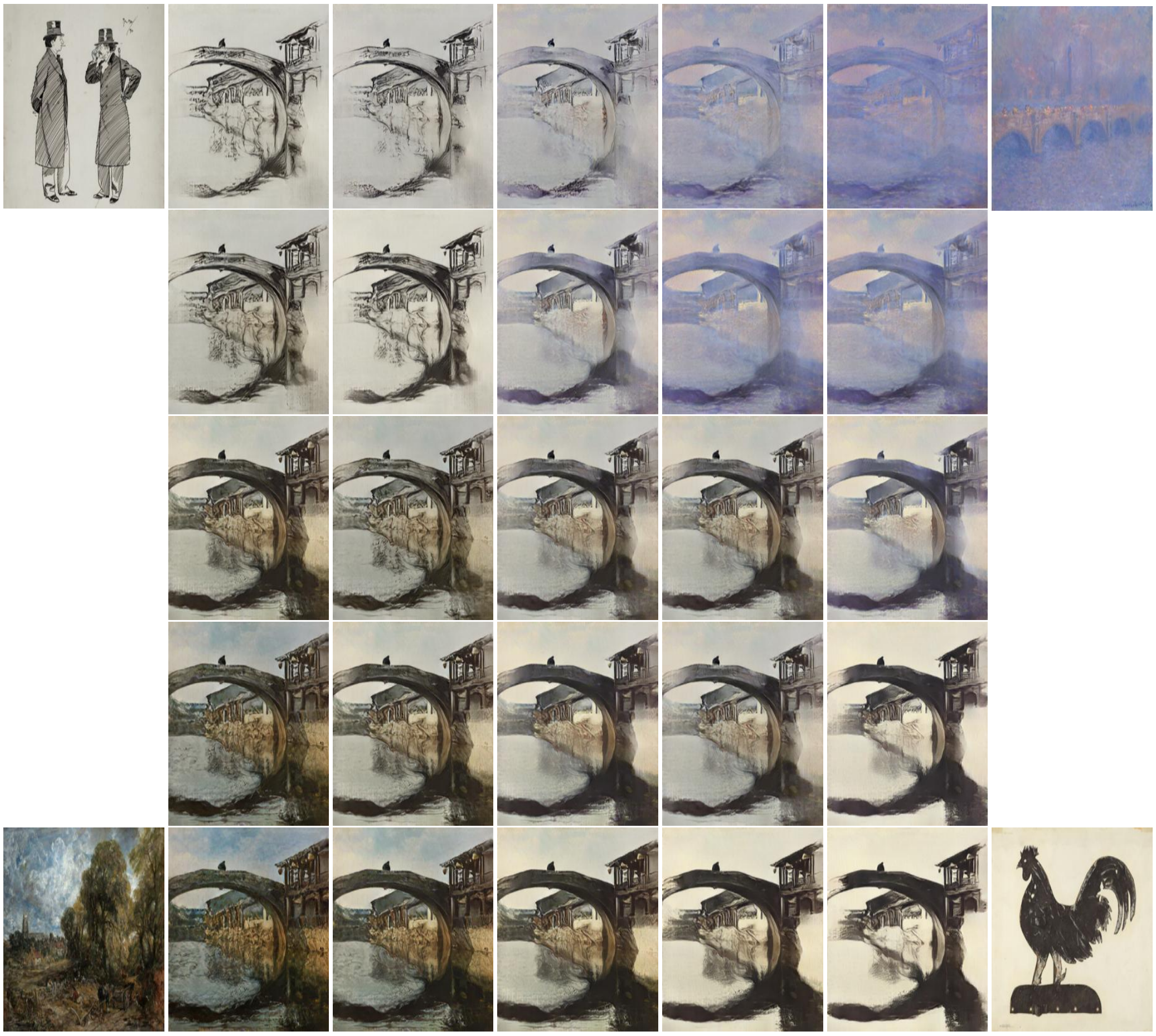}
\caption{\revision{Linear interpolation results of multiple styles. The input style images are shown in the four corners.}
%% Content image (1\textsuperscript{st} row) by courtesy of Cristian Bortes/Flickr (CC BY-NC-SA 2.0).
%% Style image credits: Master of the Small Landscapes/WikiArt (Public Domain), Alexandre Istrati/WikiArt (Fair Use).
}
\label{fig:interpolate}
\end{figure}

%%%% Figs/fig_interpolate.tex ends here %%%%

\subsection{Qualitative Evaluation}

\subsubsection{Image Style Transfer}

We first present qualitative results of our method against the selected state-of-the-art methods in Fig.~\ref{fig:SOTA}.
The comparison shows the superiority of UCAST in terms of visual quality.
AdaIN often fails to generate sharp details and introduces undesired patterns that do not exist in style images (e.g., the 4\textsuperscript{th}, 6\textsuperscript{th}, 9\textsuperscript{th} and 11\textsuperscript{th} rows).
ArtFlow sometimes generates unexpected colors or patterns in relatively smooth regions in some cases (e.g., the 2\textsuperscript{nd}, 3\textsuperscript{rd} and 8\textsuperscript{th} rows).
MCCNet can effectively preserve the input content but may fail to capture the stroke details and often generates haloing artifacts around object contours (e.g., the 2\textsuperscript{nd}, 5\textsuperscript{th}, 9\textsuperscript{th}rows).
AdaAttN cannot well capture some stroke patterns and fails to transfer important colors of the style references to the results (e.g., the 1\textsuperscript{st},  5\textsuperscript{th} and 6\textsuperscript{th} rows).
Although the generated visual effects of IEST are of high quality, the usage of second-order statistics as style representation causes color distortion (e.g., the 1\textsuperscript{st} and the 4\textsuperscript{th} row) and cannot capture the detailed stylized patterns (e.g., the 5\textsuperscript{th} and 7\textsuperscript{th} rows ).
StyleFormer cannot well capture some stroke patterns and tends to generates artifacts in the results (e.g., the 1\textsuperscript{st},  6\textsuperscript{th} and 8\textsuperscript{th} rows).
StyTr$^2$ cannot well transfer the unique style of the reference images and also tends to generates artifacts(e.g., the 1\textsuperscript{st},  3\textsuperscript{rd} and 4\textsuperscript{th} rows).
In particular, these state-of-the-art methods cannot capture the \emph{leaving blank} characteristic of Chinese painting style in the 1\textsuperscript{st} row of Fig.~\ref{fig:SOTA} and fail to generate results with a clean background.

In comparison, UCAST achieves the best stylization performance that balances characteristics of style patterns and content structures.
Instead of using second-order statistics as a global style descriptor, we use an MSP module for style encoding with the help of a DE module for effective learning of style distribution.
Thus, UCAST can flexibly represent vivid local stroke characteristics and the overall appearance while still preserving the content structure.
For instance, as shown in Figs.~\ref{fig:teaser} and \ref{fig:motivation} (the $1\textsuperscript{st}$ row) and \ref{fig:SOTA} (the $1\textsuperscript{st}$ row), UCAST successfully captures the large portion of empty regions in the style images, and it generates a stylization results which have salient objects in the center and blank space around.
As shown in Fig.~\ref{fig:SOTA}, besides commonly used oil paintings (the $2\textsuperscript{nd}$, $3\textsuperscript{rd}$ and $5\textsuperscript{th}$ rows), UCAST can also generate high quality results of line drawing (the $4\textsuperscript{th}$), cartoon (the $7\textsuperscript{th}$, $8\textsuperscript{th}$ and $9\textsuperscript{th}$ rows) , aquarelle (the $6\textsuperscript{th}$ and $11\textsuperscript{th}$ rows) , crayon drawing
 (the $10\textsuperscript{th}$ row) and color pencil drawing (the $12\textsuperscript{th}$ row).

\subsubsection{Style Interpolation}

We interpolate the feature maps among four style images with equivalent weights.
As shown in Fig.~\ref{fig:interpolate}, we can interpolate among arbitrary styles by providing the decoder with a convex mixture of feature maps converted to various styles.
Smooth intra-domain (vertically) and inter-domain (horizontally) interpolation results are obtained.

\subsection{Quantitative Evaluation}
We use the content loss~\cite{Li:2017:UST}, LPIPS~\cite{chen2021artistic}, and deception rate~\cite{Sanakoyeu:2018:SAC} and conduct two user studies to evaluate our method quantitatively.
The two user studies are online surveys that cover art/computer science students/professors and civil servants.

For content loss and LPIPS, we use a pre-trained VGG-19 and compute the average perceptual distances between the content image and the stylized image. 
The statistics are shown in Table~\ref{tab:quantitative}.
For deception rate, we train a VGG-19 network to classify 10 styles on WikiArt.
Then, the deception rate is calculated as the percentage of stylized images that are predicted by the pre-trained network as the correct target styles.
We report the deception rate for the proposed UCAST and the baseline models in the 2\textsuperscript{nd} column of Table~\ref{tab:quantitative}.
As observed, UCAST achieves the highest accuracy and surpasses other methods by a large margin.
As a reference, the mean accuracy of the network on real images of the artists from WikiArt is $78\%$.

\paragraph{User Study \uppercase\expandafter{\romannumeral1}}
We compare UCAST with seven state-of-the-art style transfer methods to evaluate which method generates results that are most favored by humans.
For each participant, 50 content-style pairs are randomly selected and the stylized results of UCAST and one of the other methods are displayed in a random order.
Then, we ask the participants to choose the image that learns the most characteristics from the style image.
Participants were told that the consistency of content and style was the primary metrics.
The style is subjective and the effectiveness of training also depends on their understanding ability.
Finally, we collect 3,800 votes from 76 participants.
We report the percentage of votes for each method in the 6\textsuperscript{th} column of Table~\ref{tab:quantitative}.
These results demonstrate that UCAST achieves the better style transfer results.
Moreover, from the statistics, we find that UCAST obtains significantly higher preferences in categories of sketch, Chinese painting, and impressionism.

\paragraph{User Study \uppercase\expandafter{\romannumeral2}}
We design a novel user study to evaluate the stylized images quantitatively, which is called the Stylized Authenticity Detection.
For each question, we show participants ten artworks of similar styles, including two to four stylized fake painting and ask them to select the synthetic ones.
Within each single question, the stylized paintings are generated by the same method.
Each participant finished 25 questions.
Finally, we collect 2000 groups of results from 80 participants and use the average precision and recall as the measurement for how likely the results will be recognized as synthetics.
We report the percentage of votes for each method in the 7\textsuperscript{th} column of Table~\ref{tab:quantitative}.
The paintings generated by UCAST have the lowest chance to be decided by people as fake paintings.
We also notice that the precision and recall of UCAST is less than $50\%$, which means that users could not tell the real ones from the fakes and tend to select more real paintings as synthetics when doing the testing.

\subsection{Video Style Transfer}

\begin{figure*}
\centering
\includegraphics[width=0.95\linewidth]{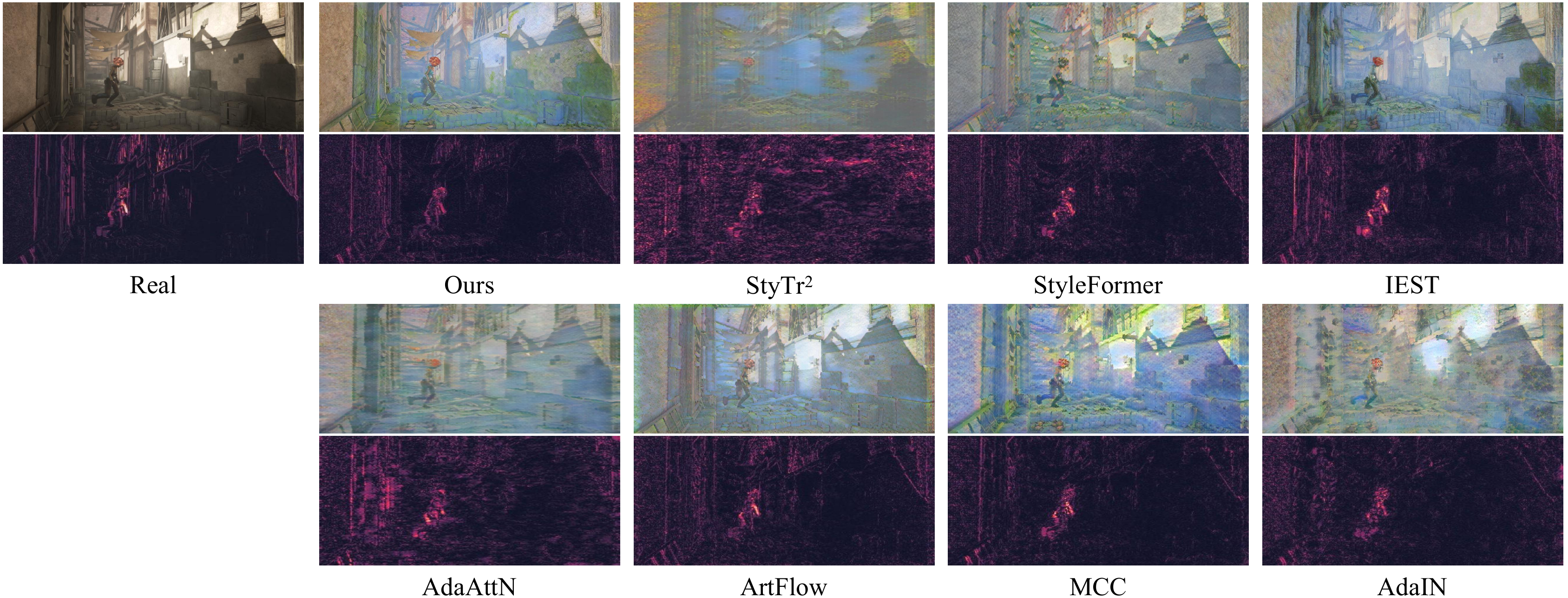}
\caption{\revision{Qualitative comparison on video style transfer.
The first column shows the input video frame and the rest of the columns show the stylization results generated with different style transfer methods.
The heat map of differences between the current frame and the previous adjacent frame are shown beneath each frame.
}}
\label{fig:video_sota}
\end{figure*}

We compare our method with seven baselines on video style transfer and show the stylization results in Fig.~\ref{fig:video_sota}.
We visualize the heat maps of differences between different frames to assess the stability and consistency of synthesized video clip.
As we can see, our approach outperforms existing style transfer methods in terms of stability and consistency by a significant margin. This can be attributed to three points: 
1) our style representation and domain distribution learning offer proper guidance to prevent the model from distorted texture patterns;
2) the cycle consistency loss enhances the consistency of synthesized video clip;
3) the added patch-wise contrastive losses offer a strong content consistency constrain which motivates the same object in a different frame to have the same stylization results.

\paragraph{Video consistency.}
We employ the widely used temporal loss ~\cite{wang:2020:consistent} to quantitatively analyze the temporal consistency of stylized videos.   
Given two adjacent frames $I_c^{t}$ and $I_c^{t-1}$ in a T-frame input clip and $I_{cs}^t$ and $I_{cs}^{t-1}$ in a T-frame rendered clip, the temporal loss is defined as:
\begin{equation}
{L}_{temporal} =  average(||O \circ (W_{I_{c}^{t-1}\rightarrow I_{c}^{t}} (I_{cs}^{t-1}) - I_{cs}^t) ||),
\end{equation}
where $O$ is an occlusion mask:
\begin{equation}
O = {|W_{I_{c}^{t-1} \rightarrow I_{c}^{t}}(I_{cs}^{t-1}) - I_{cs}^t| > 10}.
\end{equation}
The mask eliminates the negative effects brought by the inaccurate optical flow estimation. 
As shown in Table~\ref{tab:temporal_consistency}, our method achieves the best temporal consistency.

%%%% Figs/fig_video_sota.tex starts here %%%%

\begin{table*}%[h]
\centering
\caption{Quantitative evaluation of temporal consistency on $30$ rendered clips.
The best results are in \textbf{bold}.}
%\resizebox{0.9\linewidth}{!}{%
\begin{tabular}{c||c|cccccccc}
\toprule
& Ours & StyTr$^2$ & StyleFormer & IEST & AdaAttN & ArtFlow & MCCNet & AdaIN \\ \hline\hline
\textbf{Temporal Loss$\downarrow$} & \textbf{0.0322} & 0.0350 & 0.0362 & 0.0340 & 0.0349 & 0.0376 & 0.0334 & 0.0345 \\
\bottomrule
\end{tabular}
%}
%\vspace{-2mm}
\label{tab:temporal_consistency}
\end{table*}

%%% Tables/tab_video.tex ends here %%%%

% %%%% Figs/fig_temperature.tex starts here %%%%

\begin{figure*}
\centering
\includegraphics[width=0.99\linewidth]{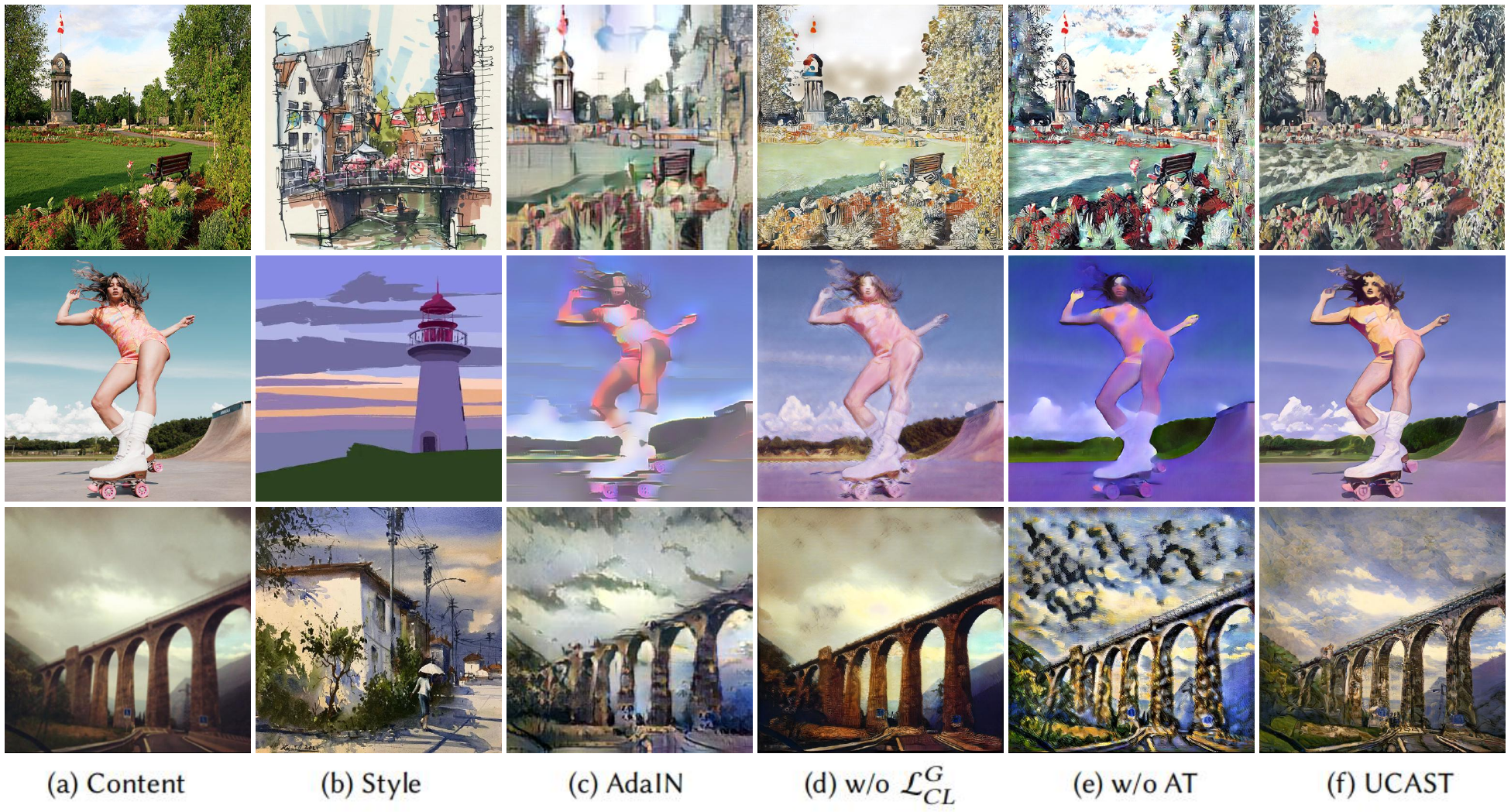}
\caption{\revision{Ablation study on adaptive contrastive learning.
From left to right: (a) content image; (b) style image; (c) AdaIN; (d) UCAST without contrastive loss; (e) UCAST without adaptive temperature ; (f) full UCAST.}
}
\label{fig:ablation_contra}
\end{figure*}

%%%% Figs/fig_temperature.tex ends here %%%%

\subsection{Ablation Study}

\paragraph{Contrastive style loss.}
We remove the contrastive style loss from Eq.~(\ref{eqn:total_loss}) to train the model.
As shown in Fig.~\ref{fig:ablation_contra_noNCE}, the model without our contrastive style loss cannot capture the color and the stroke characteristics of the style image compared with the full model.
The brushstrokes of water color in the style image almost disappear in the 1\textsuperscript{st} row.
The sharp lines and edges in the 2\textsuperscript{nd} row become smooth and murky.
The brown color of the whole image generated in the 3\textsuperscript{rd} row dose not appear in the style image.

We replace the adaptive temperature from Eq.~(\ref{eqn:loss_G_NCE2}) with constant temperature to train the model.
As shown in Fig.~\ref{fig:ablation_contra_cast}, when dealing with difficult content-style pairs, the model without our adaptive temperature tends to generate artifacts.
For instance, the black artifact appears in the sky of the 1\textsuperscript{st} row and 3\textsuperscript{rd} row.
Meanwhile, by introducing input-dependent temperature, the full UCAST can capture and transfer the unique style of cartoon.
In the 2\textsuperscript{nd} row, the sharp lines and flat color fillings in the style image are faithfully transfer to the results while the simplified model generates result with mixing style.
Meanwhile, the content details of the women's face are well preserved by the full model.
With the contrastive style loss and adaptive temperature, our full model can faithfully transfer the brushstrokes, textures, and colors from the input style image.

%%%% Figs/fig_ablation_study.tex starts here %%%%

\begin{figure*}
\centering\includegraphics[width=0.99\linewidth]{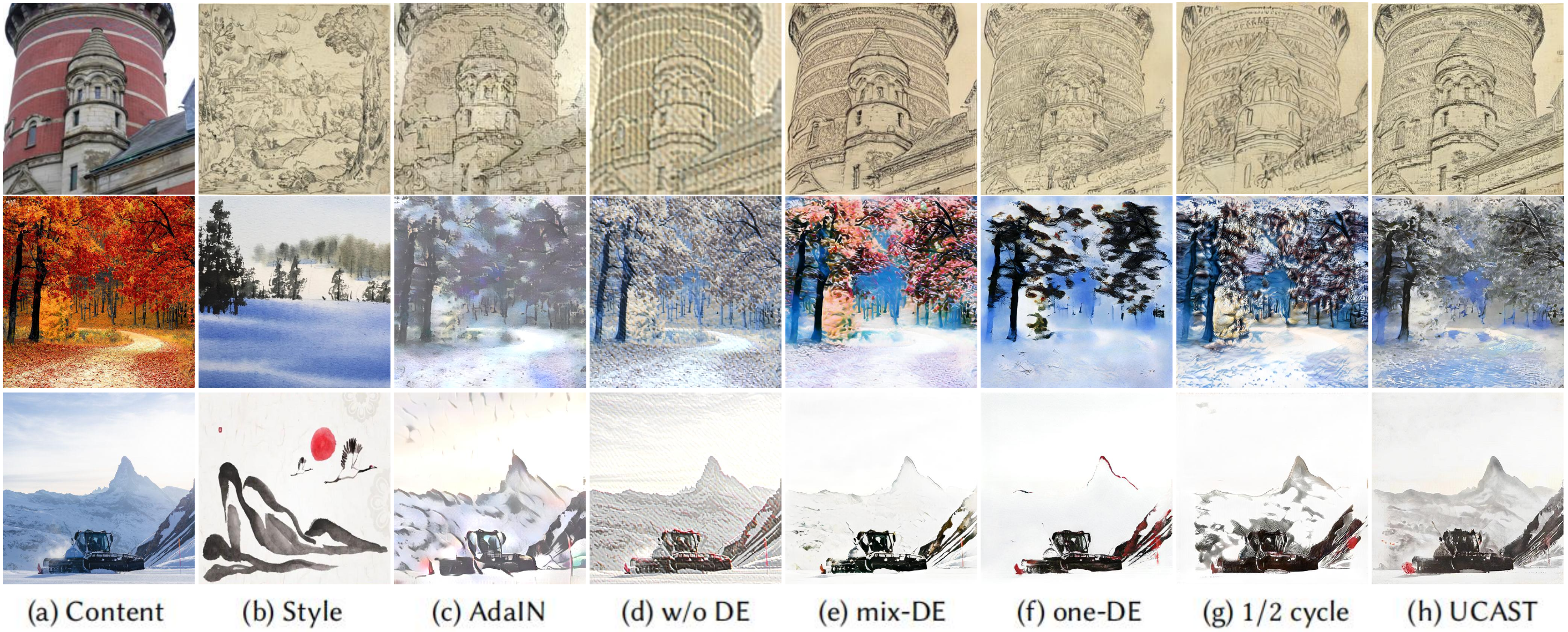}
\caption{Ablation study on domain enhancement.
From left to right: (a) content image; (b) style image; (c) AdaIN; (d) UCAST without DE; (e) UCAST using mixed DE; (f) UCAST using one DE without the realistic domain; (g) UCAST trained with asymmetric cycle consistent loss by only reconstructing the realistic images; (h) full UCAST model.
% Style image credits: Michel Ange Corneille/The Art Institute of Chicago (CC0), Steve Johnson/Pexels (CC0).
}
\label{fig:ablation_study}
\end{figure*}

% %%%% Figs/fig_ablation_study.tex ends here %%%%
\paragraph{Domain enhancement.}
Our full UCAST uses DE for realistic and artistic images separately. 
We train a simplified UCAST model without DE module.
As shown in Figs.~\ref{fig:ablation_study_no_domain} and~\ref{fig:ablation_study_full}, the color of the style images are faithfully transferred, but the generated images do not appear like real paintings.
We train a simplified UCAST model using one discriminator that mixes realistic and artistic images together (mix-DE).
As shown in Fig.~\ref{fig:ablation_study_one_domain}, the results generated by mix-DE model are acceptable, but the stroke details in the generated images are weaker than the ones by the full UCAST model.
This fact is due to the existence of a significant gap between the artistic and realistic image domains. 
We further abandon all images from realistic domain for ablation (one-DE).
As shown in Fig.~\ref{fig:ablation_study_art_domain}, the results generated by one-DE model lack details.

To better evaluate the improvement of the contrastive style loss on the style transfer task, we exclude the latent promotion of cycle consistency loss from network training.
The reason is that the reconstruction process of artistic image may imply style information.
We train UCAST with an asymmetric cycle consistent loss, which only reconstructs the realistic images.
The decoder of the style transfer network is unaffected by the reconstruction of the artistic image.
As shown in Fig.~\ref{fig:ablation_study_cyc}, removing realistic image reconstruction will lead to slightly degraded stylization results.

%%% 4-Experiments.tex ends here %%%%
%%%% 5-Conclusion.tex starts here %%%%

\section{Conclusion and Future Work}
In this work, we present a novel unified framework, namely UCAST, for the task of arbitrary image style transfer. 
Instead of relying on second-order metrics such as Gram matrix or mean/variance of deep features, we use image features directly by introducing an MSP module for style encoding. 
We develop a parallel contrastive learning scheme to leverage the available multi-style information in the existing collection of artwork and help train the MSP module and the generative style transfer network.
We propose an adaptive contrastive learning for style transfer process implemented by a dual input-dependent temperature.
We further propose a DE scheme to effectively model the distribution of realistic and artistic image domains. 
Extensive experimental results demonstrate that our proposed UCAST method is effective for various generative backbones and achieves superior arbitrary style transfer results compared with state-of-the-art approaches. 
In the future, we plan to improve the contrastive style learning process by considering artist and category information.

%%%% 5-Conclusion.tex ends here %%%%

\bibliographystyle{ACM-Reference-Format}
\balance
\bibliography{UAST}

\end{document}